\documentclass[10pt,twocolumn,letterpaper]{article}

\usepackage{cvpr}
\usepackage{times}
\usepackage{epsfig}
\usepackage{graphicx}
\usepackage{amsmath}
\usepackage{amssymb,enumitem}

\usepackage{multirow}
\usepackage{tabls}
\usepackage{wrapfig}
\usepackage{hyperref}
\usepackage{amsmath,amssymb} 
\usepackage{amsmath,amssymb}
 \usepackage[ruled,vlined,linesnumbered,resetcount]{algorithm2e}
  \usepackage{booktabs}
 \usepackage[linesnumbered]{algorithm2e}
\usepackage{algorithmic}
\usepackage{caption}
\usepackage{subfigure}
\renewcommand{\algorithmicrequire}{ \textbf{Input:}} 
\renewcommand{\algorithmicensure}{ \textbf{Output:}} 
\usepackage[font=small,labelfont=bf]{caption}
   \usepackage{graphicx}
 \usepackage{booktabs}

\usepackage{epstopdf}
\usepackage{graphicx}
\usepackage{subfigure}
\usepackage{enumitem}
\usepackage{multirow}
\usepackage{tabls}
\usepackage{wrapfig}
\usepackage{hyperref}
\usepackage{amsmath,amssymb} 
\usepackage{amsmath,amssymb}
 \usepackage[ruled,vlined,linesnumbered,resetcount]{algorithm2e}
  \usepackage{booktabs}
 \usepackage[linesnumbered]{algorithm2e}
\usepackage{algorithmic}
\usepackage{caption}
\usepackage{subfigure}
\renewcommand{\algorithmicrequire}{ \textbf{Input:}} 
\renewcommand{\algorithmicensure}{ \textbf{Output:}} 
\usepackage[font=small,labelfont=bf]{caption}
   \usepackage{graphicx}
 \usepackage{booktabs}
\usepackage{epstopdf}
\usepackage{graphicx}
\usepackage{subfigure}
\usepackage{enumitem}

\cvprfinalcopy 

\def\cvprPaperID{49} 

\ifcvprfinal\fi
\begin{document}

\title{Densely Connected Pyramid Dehazing Network}

\author{He  Zhang \qquad \qquad \qquad Vishal M. Patel \\
Department of Electrical and Computer Engineering\\Rutgers University, Piscataway, NJ 08854
\\
{\tt\small {\{he.zhang92,vishal.m.patel\}}@rutgers.edu}
}

\setlength\abovedisplayskip{0pt}
    \setlength\belowdisplayskip{0pt}

\maketitle

\begin{abstract}
We propose a new end-to-end single image dehazing method, called Densely Connected Pyramid Dehazing Network (DCPDN), which can jointly learn the transmission map, atmospheric light and dehazing all together. The end-to-end learning is achieved by  directly embedding the atmospheric scattering model into the network, thereby ensuring that the proposed method strictly follows the physics-driven scattering model for dehazing.  Inspired by the dense network that can maximize the information flow along features from different levels, we propose a new edge-preserving densely connected encoder-decoder structure with multi-level pyramid pooling module for estimating the transmission map. This network is optimized using a newly introduced edge-preserving loss function. To further incorporate the mutual structural information between the estimated transmission map and the dehazed result, we propose a joint-discriminator based on generative adversarial network framework to decide whether the corresponding dehazed image and the estimated transmission map are real or fake. An ablation study is conducted to demonstrate the effectiveness of each module evaluated at both estimated transmission map and dehazed result. Extensive  experiments    demonstrate   that   the proposed  method  achieves  significant  improvements  over  the state-of-the-art  methods. Code will be made available at: https://github.com/hezhangsprinter
\end{abstract}

\section{Introduction}
Under severe hazy conditions, floating particles in the atmosphere  such as dusk and smoke greatly absorb and scatter the light, resulting in degradations in the image quality. These degradations in turn may affect the performance of many computer vision systems such as classification and detection.   To overcome the degradations caused by haze, image and video-based haze removal algorithms have been proposed in the literature  \cite{dehaze_2016_eccv,dehaze_2016_tip,dehaze_2014_tang,dehaze_2016_dana,dehaz_2009_dark,dehaze_depth,dehaze_multi_image,dehaze_2017_joint,dehaze_iccv17,dehaze_fast15,Fattal_2008_siggraph,dehaze_2014acm,dehaze_2008,ren2018gated}.  
\begin{figure}[t]
	\centering
	\begin{minipage}{.23\textwidth}
		\centering
		\includegraphics[width=4.05cm, height=3.2cm]{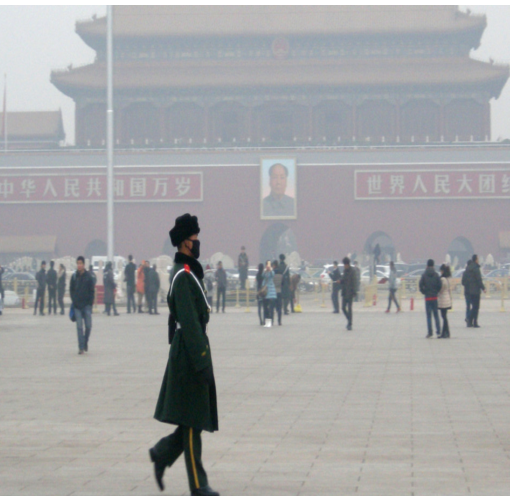}
		\captionsetup{labelformat=empty}
		\captionsetup{justification=centering}
	\end{minipage}
	\begin{minipage}{.23\textwidth}
		\centering
		\includegraphics[width=4.05cm, height=3.2cm]{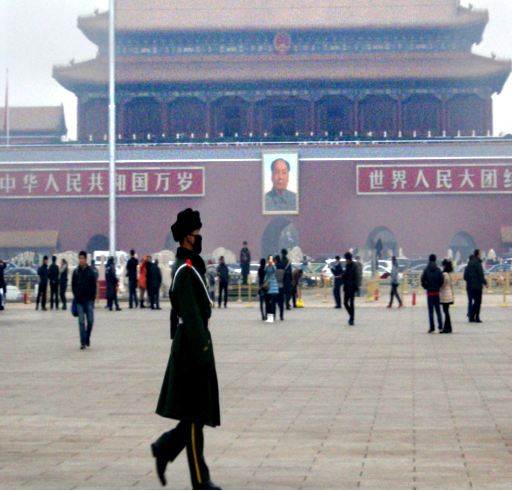}
		\captionsetup{labelformat=empty}
		\captionsetup{justification=centering}
	\end{minipage}	
	\vskip -4pt\caption{Sample image dehazing result using the proposed DCPDN method. Left: Input hazy image.  Right: Dehazed result.} \label{fig:over}
	\vskip -18pt
\end{figure}

The  image  degradation (atmospheric scattering model)  due  to  the  presence  of  haze  is mathematically formulated as
\begin{equation}
{I}({z})={J(z)}t({z})+{A(z)}(1-t({z})),
\label{eq: intro}
\end{equation}
where ${I}$ is the observed hazy image, $J$ is the true scene radiance, ${A}$ is the global atmospheric light, indicating the intensity of the ambient light, $t$ is the transmission map and $z$ is the pixel location.  Transmission map is the distance-dependent factor that affects the fraction of light that reaches the camera sensor. When the atmospheric light $A$ is homogeneous, the transmission map can be expressed as $t{(z)}=e^{-\beta d({z})}$, where $\beta$ represents  attenuation coefficient of the atmosphere and $d$ is the scene depth.    In single image dehazing, given $I$, the goal is to estimate $J$.

It can be observed from Eq.~\ref{eq: intro} that there exists two important aspects in the dehazing process: \emph{(1) accurate estimation of transmission map,} and \emph{(2) accurate estimation of atmospheric light}.  Apart from  several works that focus on estimating the atmospheric light \cite{dehaze_iccp217,dehaze_iccp14_light}, most of the other algorithms concentrate more on the accurate estimation of the transmission map and they leverage empirical rule in estimating the atmospheric light \cite{dehaz_2009_dark,dehaze_2013_meng,dehaze_2016_eccv,dehaze_tan_2008}. This is mainly due to the common belief that good estimation of transmission map will lead to better dehazing.  These methods can be broadly divided into two main groups: prior-based methods and learning-based methods.  Prior-based methods often leverage different priors in characterizing the transmission map such as dark-channel prior \cite{dehaz_2009_dark},  contrast color-lines \cite{dehaze_2014acm} and haze-line prior \cite{dehaze_2016_dana}, while learning-based methods, such as those based on convolutional neural networks (CNNs), attempt to learn the transmission map directly from the training data \cite{dehaze_2014_tang,dehaze_2016_eccv,dehaze_2016_tip,dehaze_2017_joint,dehaze_iccv17}.   Once the transmission map and the atmospheric light are estimated, the dehazed image can be recovered as follows 
\begin{equation}
{\hat{J}(z)}=\frac{{I(z)}-{\hat{A}(z)}(1-\hat{t}({z}))}{\hat{t}({z})}.
\label{eq: dehaze1}
\end{equation}

\begin{figure}[t]

\centering
		\includegraphics[width=0.5\textwidth]{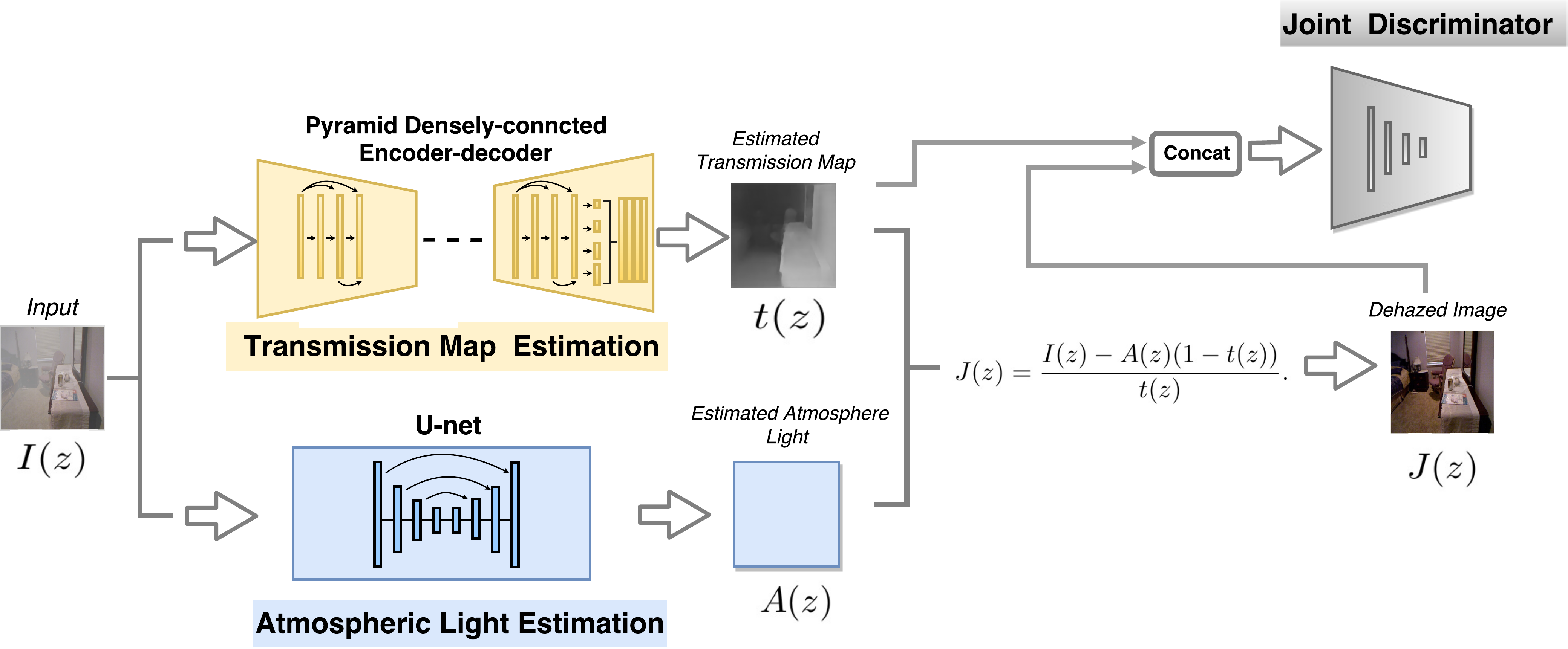}
   \caption{An overview of the proposed DCPDN image dehazing method. DCPDN consists of four modules: 1. Pyramid densely connected transmission map estimation net. 2. Atmospheric light estimation net. 3. Dehazing via Eq\ref{eq: dehaze1}. 4. Joint discriminator.  We first estimate the transmission map using the proposed pyramid densely-connected transmission estimation net, followed by prediction of atmospheric light using the U-net structure. Finally, using the estimated transmission map and the atmospheric light we estimate the dehazed image via Eq.~\ref{eq: dehaze1}.}
\label{fig:overview}
	\vskip -6pt
\end{figure}

Though tremendous improvements have been made by the learning-based methods, several factors hinder the performance of these methods and the results are far from optimal.  This is mainly because: \emph{1. Inaccuracies in the estimation of transmission map translates to low quality dehazed result. 2. Existing methods do not leverage end-to-end learning and are unable to capture the inherent relation among transmission map, atmospheric light and dehazed image. The disjoint optimization may hinder the overall dehazing performance.} Most recently, a method was proposed  in  \cite{dehaze_iccv17} to jointly optimize the whole dehazing network.  This was achieved by leveraging a linear transformation to embed both the transmission map and the atmospheric light into one variable and then learning a light-weight CNN to recover the clean image.

In this paper, we take a different approach in addressing the end-to-end learning for image dehazing.   In particular, we propose a new image dehazing architecture, called Densely Connected Pyramid Dehazing Network (DCPDN), that can be jointly optimized to estimate transmission map, atmospheric light and also image dehazing simultaneously by following the image degradation model Eq.~\ref{eq: intro} (see Fig.~\ref{fig:overview}).   In other words, 
the end-to-end learning is  achieved by embedding Eq.~\ref{eq: intro} directly into the network via the math operation modules provided by the deep learning framework. However, training such a complex network (with three different tasks) is very challenging. To ease the training process and accelerate the network convergence, we leverage a stage-wise learning technique in which we first progressively optimize each part of the network and then jointly optimize the entire network.  To make sure that the estimated transmission map preserves sharp edges and avoids halo artifacts when dehazing,  a new edge-preserving loss function is proposed in this paper based on the observation that gradient operators and first several layers of a CNN structure can function as edge extractors.   Furthermore, a densely connected encoder-decoder network with multi-level pooling modules is proposed  to leverage features from different levels for estimating the transmission map. To exploit the structural relationship between the transmission map and the dehazed image, a joint discriminator-based generative adversarial network (GAN) is proposed. The joint discriminator distinguishes whether a pair of estimated transmission map and dehazed image is a real or fake pair. To guarantee that the atmospheric light can also be optimized within the whole structure, a U-net \cite{unet} is adopted to estimate the homogeneous atmospheric light map.  Shown in Fig.~\ref{fig:over} is a sample dehazed image using the proposed method.

This paper makes the following contributions:
\begin{itemize}[noitemsep]
  \item A novel end-to-end jointly optimizable dehazing network is proposed. This is enabled by embedding Eq.~\ref{eq: intro} directly into the optimization framework via math operation modules. Thus, it allows the network to estimate the transmission map, atmospheric light and dehazed image jointly. The entire network is trained by a stage-wise learning method.
  \item An edge-preserving pyramid densely connected encoder-decoder network is proposed for  accurately estimating the transmission map.  Further, it is optimized via a newly proposed edge-preserving loss function.
  \item As the structure of the estimated transmission map and the dehazed image are highly correlated,  we leverage a joint discriminator within the GAN framework to determine whether the paired samples (i.e. transmission map and dehazed image) are from the data distribution or not. 
  \item   Extensive experiments are conducted on  two synthetic datasets and one real-world image dataset.  In addition, comparisons are  performed  against  several  recent  state-of-the-art approaches. Furthermore, an ablation study is conducted to demonstrate the improvements obtained by different modules in the proposed network.

\end{itemize}
\section{Related Work}
\noindent {\bf{Single Image Dehazing.}}    Single image dehazing is a highly ill-posed problem.   Various handcrafted prior-based and learning-based methods have been developed to tackle this problem.\\  
\emph{Handcrafted Prior-based}: Fattal~\cite{dehaze_2008} proposed a physically-grounded method by estimating the albedo of the scene.  As the images captured from the hazy conditions always lack color contrast, Tan \cite{dehaze_tan_2008} \emph{et al}.  proposed a patch-based contrast-maximization method.  In  \cite{dehaze_2009_kratz}, Kratz and Nishino  proposed a factorial  MRF model to estimate the albedo and depths filed. Inspired by the observations that outdoor objects in clear weather have at least one color channel that is significantly dark, He. \emph{et al.} in \cite{dehaz_2009_dark} proposed a dark-channel model to estimate the transmission map.  More recently, Fattal \cite{dehaze_2014acm} proposed a color-line method based on the observation that small image patches typically exhibit a one-dimensional distribution in the RGB color space. Similarly, Berman \emph{et al.} \cite{dehaze_2016_dana} proposed a non-local patch prior to characterize the clean images.\\
\emph{Learning-based:} Unlike some of the above mentioned methods that use different priors to estimate the transmission map,  Cai \emph{et al.}  \cite{dehaze_2016_tip} introduce an end-to-end CNN network for estimating the transmission with a novel BReLU unit. More recently, Ren \emph{et al.} \cite{dehaze_2016_eccv} proposed a multi-scale deep neural network to estimate the transmission map. One of the limitations of these methods is that they limit their capabilities by only considering the transmission map in their CNN frameworks. To address this issue, Li. \emph{et al} \cite{dehaze_iccv17} proposed an  all-in-one dehazing network, where a linear transformation is leveraged to encode the transmission map and the atmospheric light into one variable. Most recently, several  benchmark datasets of both synthetic and real-world hazy images for dehazing problems are introduced to the community \cite{Zhang:HazeRD:ICIP17b,2017arXiv171204143L}.\\

\noindent {\bf{Generative Adversarial Networks (GANs).}} The notion of GAN was first proposed by Goodfellow \textit{et al.} in \cite{gan} to synthesize realistic images by effectively learning the distribution of the training images via  a game theoretic min-max optimization framework. The success of GANs in synthesizing realistic images has led researchers to explore the adversarial loss for various low-level vision applications such as text-to-image synthesis\cite{reed2016generative,gan_stack_zhang,zizhao_cvpr_text,2018arXiv180100077D}, image-image translation \cite{pix2pix2016,gan_domain_liu,wang2017high} and other applications \cite{GAN_sisr,derain_2017_zhang,zizhao_cvpr_medical,yizhe_cvpr18,lidan_fg,sindagi2017generating,wang2018STCGAN,xi_peng_gan}. Inspired by the success of these methods in generating high-quality images with fine details, we propose a joint discriminator-based GAN to refine the estimated transmission map and dehazed image.

\section{Proposed Method}
The proposed DCPDN network architecture is illustrated in Fig.~\ref{fig:overview} which consists of the following four modules: 1) Pyramid densely connected transmission map estimation net, 2) Atmosphere light estimation net, 3) Dehazing via Eq.~\ref{eq: dehaze1}, and 4) Joint discriminator.  In what follows, we explain these modules in detail.\\

\noindent {\bf{Pyramid Densely Connected Transmission Map Estimation Network.}} Inspired by the previous methods that use multi-level features for estimating the transmission map \cite{dehaze_2016_eccv,dehaze_2016_tip,dehaze_2014_tang,dehaze_mul_fusion,dehaze_iccv17},  we propose a densely connected encoder-decoder structure that makes use of the features from multiple layers of a CNN, where the dense block is used as the basic structure. The reason to use  dense block lies in that it can maximize the information flow along those features and guarantee better convergence via connecting all layers.    In addition,  a multi-level pyramid pooling module is adopted to  refine the learned features by considering the `global' structural information into the optimization \cite{psp_net}.  To leverage the pre-defined weights of the dense-net \cite{dense_net}, we adopt the first \emph{Conv} layer and the first three \emph{Dense-Blocks} with their corresponding down-sampling operations \emph{Transition-Blocks} from a pre-trained dense-net121  as our encoder structure. The feature size at end of the encoding part is 1/32 of the input size. To reconstruct the transmission map into the original resolution, we stack five dense blocks with the refined up-sampling \emph{Transition-Blocks} \cite{dense_fully,zhu2017densenet} as the decoding module. In addition, concatenations are employed with the features corresponding to the same dimension.

\begin{figure}[htp!]
\centering
		\includegraphics[width=0.5\textwidth]{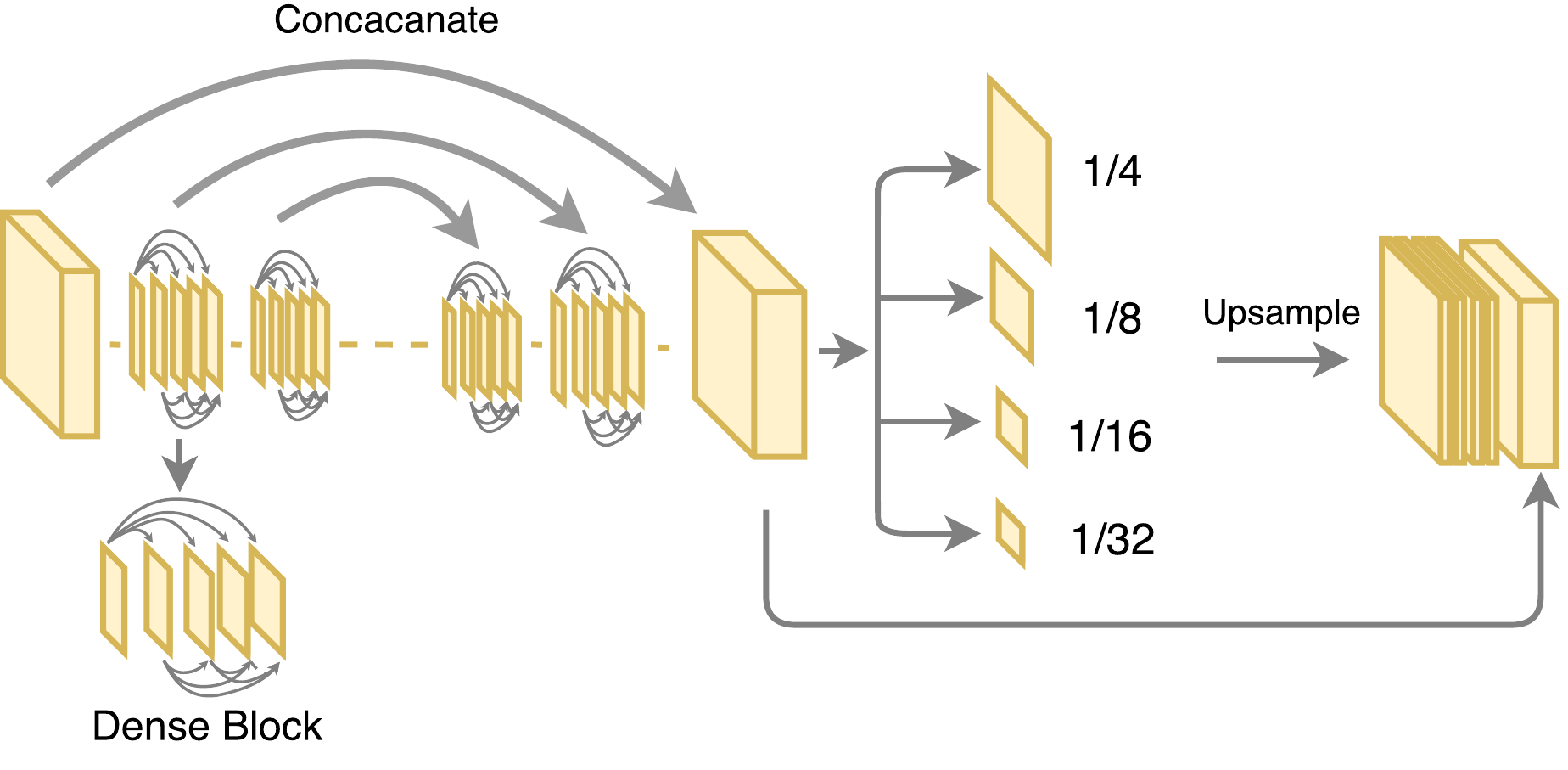}
 \vskip -5pt  \caption{An overview of the proposed pyramid densely connected transmission map estimation network.}
\label{fig:overview_tran}
\end{figure}

Even though the proposed densely connected encoder-decoder structure combines different features within the network, the result from just densely connected structure still lack of  the `global' structural information of objects with different scales. One possible reason is that the features from different scales are not used to directly estimate the final transmission map.  To efficiently address this issue,  a multi-level pyramid pooling block is adopted to make sure that features from different scales are embedded in the final result. This is inspired by the use of global context information in classification and segmentation tasks \cite{psp_net,hang_2018_CVPR,Spatial_pymaid}. Rather than taking very large pooling size to capture more global context information between different objects \cite{psp_net}, more `local' information to characterize the `global' structure of each object is needed.  Hence, a four-level pooling operation with pooling sizes 1/32, 1/16, 1/8 and 1/4 is adopted. Then, all four level features are up-sampling to original feature size and are concatenated back with the original feature before the final estimation.  Fig~\ref{fig:overview_tran} gives an overview of the proposed pyramid densely connected transmission map estimation network.\\

\noindent {\bf{Atmospheric Light Estimation Network.}} Following the image degradation model  Eq.;~\ref{eq: intro}, we assume that the atmospheric light map $A$ is homogeneous \cite{dehaz_2009_dark,dehaze_2016_tip}. Similar to previous works, the predicted  atmospheric light $A$ is uniform for a given image. In other words,  the predicted $A$ is a 2D-map, where each pixel $A(z)$ has the same value (eg. $A(z)=c, \;c\;\; \text{is a constant}$). As a result, the ground truth $A$ is of the same feature size as the input image and the pixels in $A$ are filled with the same value.  To estimate the atmospheric light, we adopt a  8-block U-net \cite{unet} structure, where the encoder is composed of four \emph{Conv-BN-Relu} blocks and the decoder is composed of symmetric \emph{Dconv-BN-Relu} block \footnote{Con: Convolution, BN: Batch-normalization \cite{batch_norm} and Dconv: Deconvolution (transpose convolution).}.\\ 

\noindent {\bf{Dehazing via Eq.~\ref{eq: dehaze1}.}} To bridge the relation among the transmission map, the atmospheric light and the dehazed image and to make sure that the whole network structure is jointly optimized for all three tasks, we directly embed \eqref{eq: dehaze1} into the overall optimization framework.  An overview of the entire DCPDN structure is shown in Fig~\ref{fig:over}.\\

\begin{figure}[t]
	\centering
	\begin{minipage}{.20\textwidth}
		\centering
		\includegraphics[width=1\textwidth]{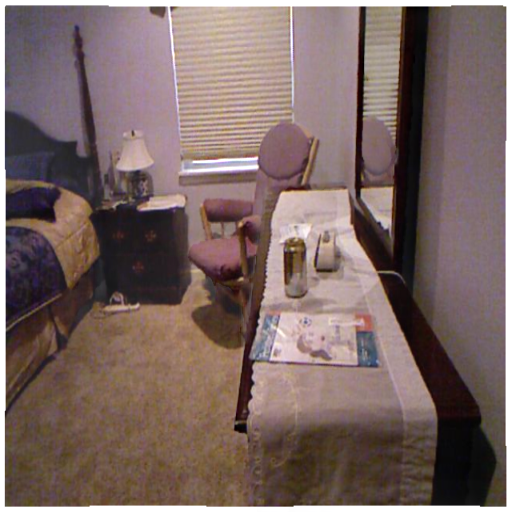}
		\captionsetup{labelformat=empty}
		\captionsetup{justification=centering}
	\end{minipage}
	\begin{minipage}{.20\textwidth}
		\centering
		\includegraphics[width=1\textwidth]{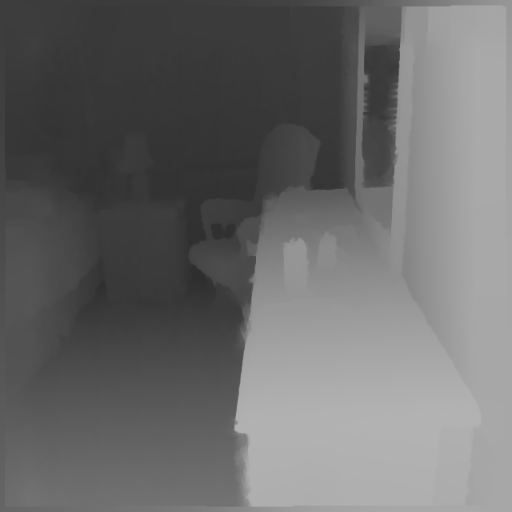}
		\captionsetup{labelformat=empty}
		\captionsetup{justification=centering}
	\end{minipage}	
	\vskip -8pt\caption{Left: a dehazed image. Right: The transmission map used to produce a hazy image from which the dehazed image on the left was obtained.} \label{fig:sample}
	\vskip -6pt
\end{figure}
\normalsize
\subsection{Joint Discriminator Learning} Let $G_t$ and $G_d$ denote the networks that generate the  transmission map and the dehazed result, respectively.  To refine the output and to make sure that the estimated  transmission map $G_t(I)$  and the dehazed image $G_d(I)$ are indistinguishable from their corresponding ground truths $t$ and $J$, respectively, we make use of a GAN~\cite{gan} with novel  joint discriminator. 

It can be observed from \eqref{eq: intro} and also Fig.~\ref{fig:sample} that the structural information between the estimated transmission map $\hat{t}=G_t(I)$ and the dehazed image $\hat{J}$ are highly correlated. Hence, in order to leverage the dependency in structural information between these two modalities, we introduce a joint discriminator to learn a joint distribution to decide whether the corresponding pairs (\emph{transmission map},  \emph{dehazed image}) are real or fake. By leveraging the joint distribution optimization, the structural correlation between them can be better exploited.  Similar to previous works, the predicted  air-light $A$ is uniform for a given image. In other words,  the predicted air-light $A$ is a 2D-map, where each pixel $A(z)$ has the same value (eg. $A(z)=c, \;c\;\; \text{is a constant}$).  We propose the following joint-discriminator based optimization
\begin{equation}\label{eq:GAN_jd}
\begin{split}
\min_{G_t, G_d} \max_{D_joint} \quad & \mathbb E _{{I}\sim p_{data({I})}}[\log (1- D_joint(G_t({I})))]+\\
& \mathbb E _{{I}\sim p_{data({I})}}[\log (1- D_joint( G_d({I})))]+\\
&\mathbb E _{{{t},{J}}\sim p_{data({t},{J})}}[\log D_{joint}({{t},{J}}))].
\end{split}
 \end{equation}

In practice, we concatenate the dehazed image with the estimated transmission map as a pair sample and then feed it into the discriminator.

\subsection{Edge-preserving Loss}
\begin{figure}[t]
	\centering
	\includegraphics[height=1.55cm]{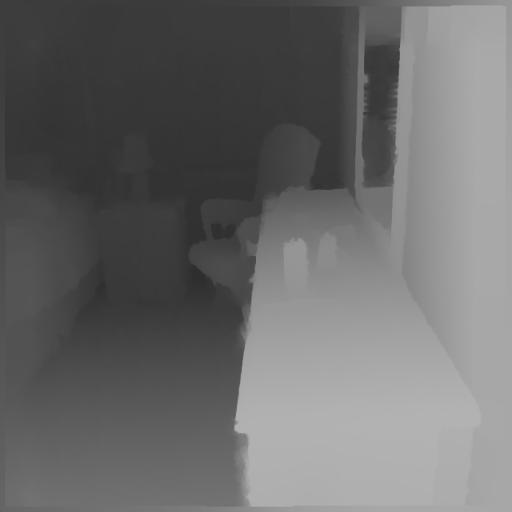}
	\includegraphics[height=1.55cm]{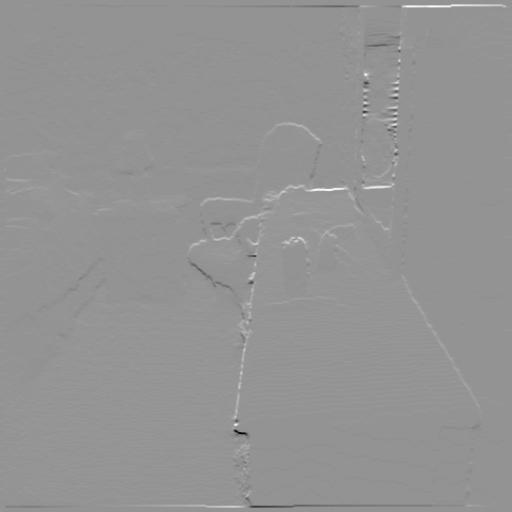}
	\includegraphics[height=1.55cm]{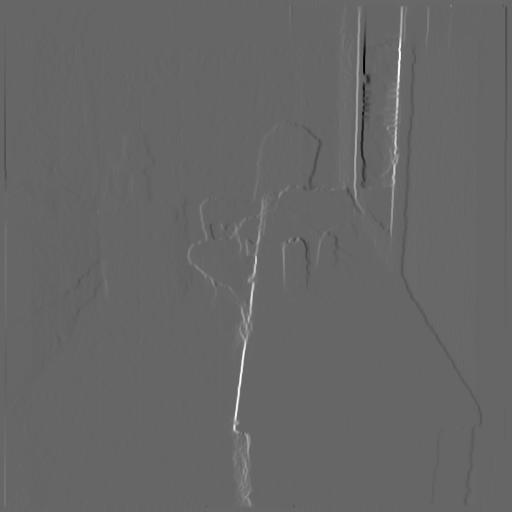}
	\includegraphics[height=1.55cm]{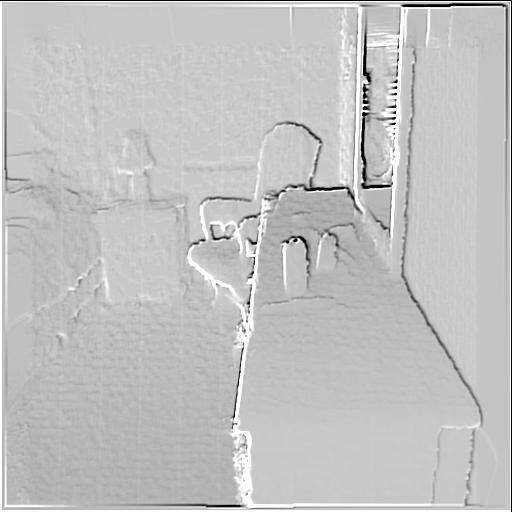}
	\includegraphics[height=1.55cm]{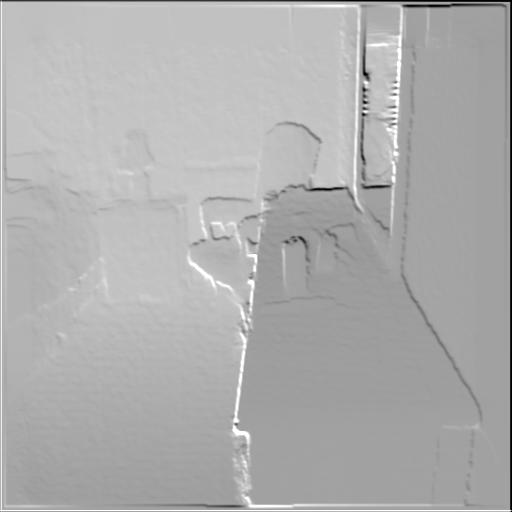}\\
	(a)\hskip30pt(b)\hskip40pt(c)\hskip40pt(d)\hskip40pt(e)\hskip40pt
	\vskip -8pt\caption{Feature visualization for gradient operator and low-level features. (a) Input transmission map. (b) Horizontal gradient output.  (c) Vertical gradient output. (d) and (e) are visualization of two feature maps from relu1\_2 of VGG-16 \cite{vgg}.}
	\label{fig:edge_vis}
	\vskip -6pt
\end{figure}

It is commonly acknowledged that the Euclidean loss ($L2$ loss) tends to blur the final result. Hence, inaccurate estimation of the transmission map with just the $L2$ loss may result in the loss of details, leading to the halo artifacts in the dehazed image \cite{dehaze_vis_edge}. To efficiently address this issue, a new edge-preserving loss is proposed,  which is motivated by the following two observations. 1) Edges corresponds to the discontinuities in the image intensities, hence it can be characterized by the image gradients. 2) It is known that low-level features such as edges and contours can be captured in the shallow (first several) layers of a CNN structure \cite{cnn_vis}. In other words, the first few layers function as an edge detector in a deep network. For example, if the transmission map is fed into a pre-defined VGG-16 \cite{vgg} model and then certain features from the output of layer relu1\_2  are visualized, it can be  clearly observed that the edge information being preserved in the corresponding feature maps (see Fig.~\ref{fig:edge_vis}).

\begin{figure*}[t]
	\centering
	\vskip -6pt
	\begin{minipage}{.16\textwidth}
		\centering
	\caption*{\emph{SSIM:0.9272}}
	\vskip -12pt
	\includegraphics[height=2.75cm]{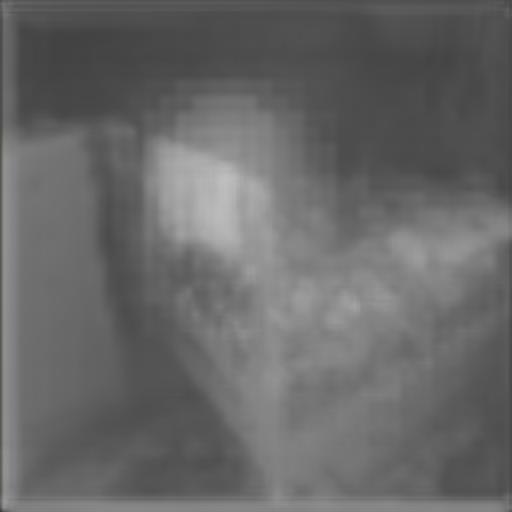}
		\captionsetup{labelformat=empty}
		\captionsetup{justification=centering}
	\end{minipage} 
	\begin{minipage}{.16\textwidth}
		\centering
	\caption*{\emph{SSIM:0.9524}}
	\vskip -12pt
	\includegraphics[height=2.75cm]{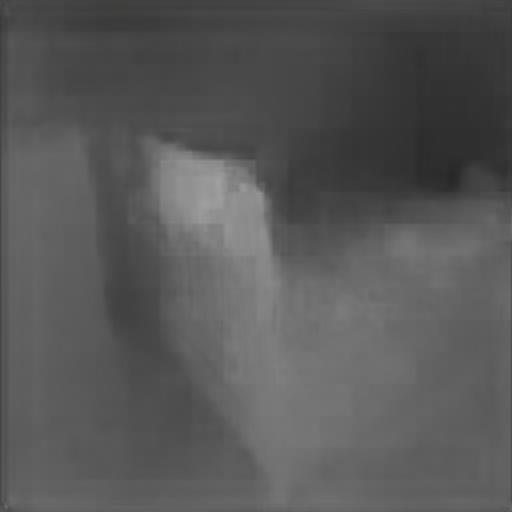}
		\captionsetup{labelformat=empty}
		\captionsetup{justification=centering}
	\end{minipage}
	\begin{minipage}{.16\textwidth}
		\centering
	\caption*{\emph{SSIM:0.9671}}
	\vskip -12pt
	\includegraphics[height=2.75cm]{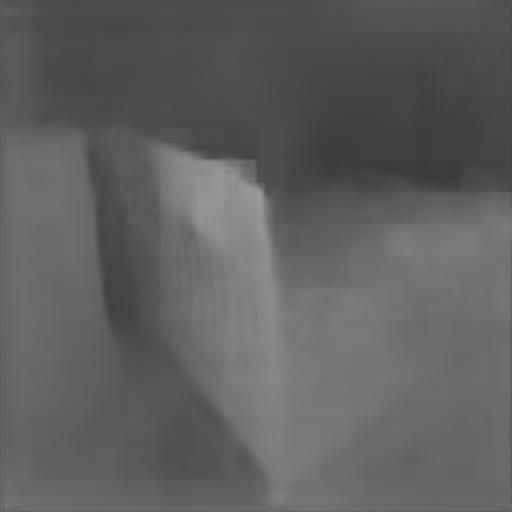}
		\captionsetup{labelformat=empty}
		\captionsetup{justification=centering}
	\end{minipage} 
	\begin{minipage}{.16\textwidth}
		\centering
	\caption*{SSIM:0.9703}
	\vskip -12pt
	\includegraphics[height=2.75cm]{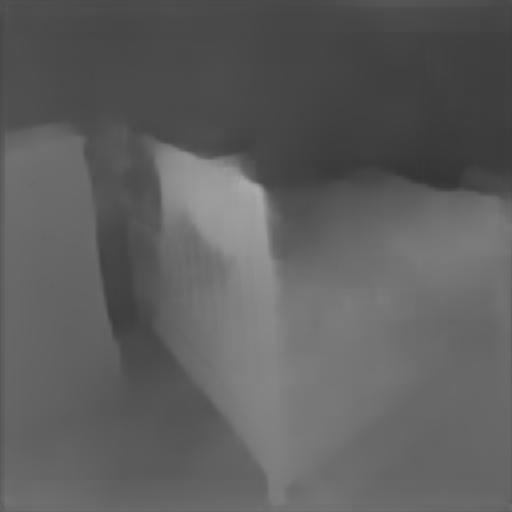}
		\captionsetup{labelformat=empty}
		\captionsetup{justification=centering}
	\end{minipage}
	\begin{minipage}{.16\textwidth}
		\centering
	\caption*{\emph{SSIM:\textbf{0.9735}}}
	\vskip -12pt
	\includegraphics[height=2.75cm]{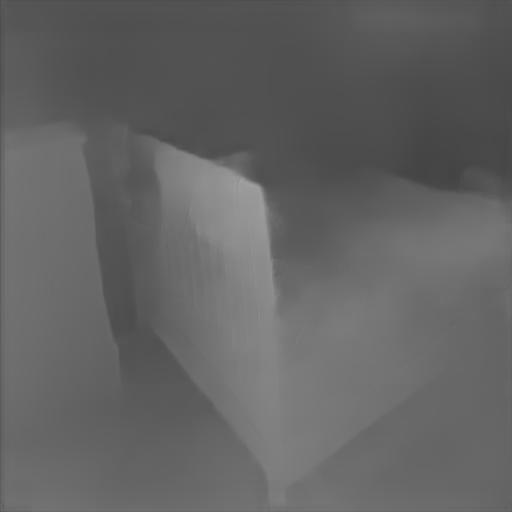}
		\captionsetup{labelformat=empty}
		\captionsetup{justification=centering}
	\end{minipage}
	\begin{minipage}{.16\textwidth}
		\centering
	\caption*{\emph{SSIM:1}}
	\vskip -12pt
	\includegraphics[height=2.75cm]{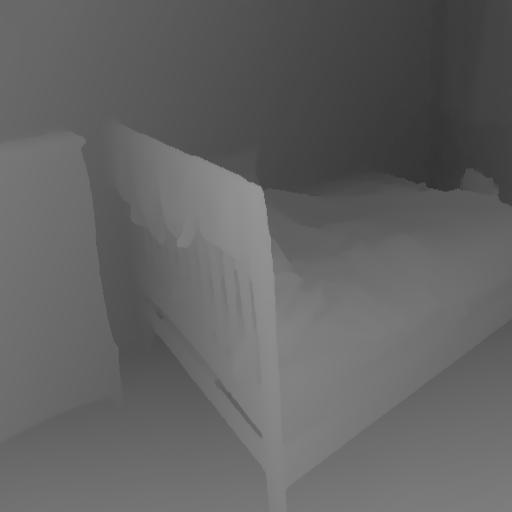}\\
		\captionsetup{labelformat=empty}
		\captionsetup{justification=centering}
	\end{minipage}		\vskip+6pt
	\begin{minipage}{.16\textwidth}
		\centering
	\caption*{\emph{SSIM:0.8882}}
	\vskip -12pt
	\includegraphics[height=2.75cm]{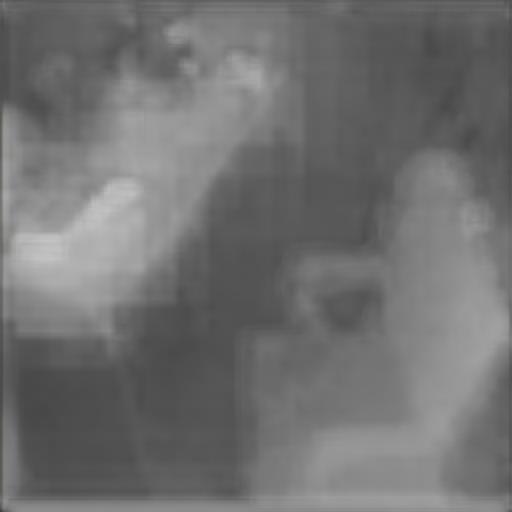}
		\captionsetup{labelformat=empty}
		\captionsetup{justification=centering}
		\caption*{(a)}
	\end{minipage} 
	\begin{minipage}{.16\textwidth}
		\centering
	\caption*{\emph{SSIM:0.9119}}
	\vskip -12pt
	\includegraphics[height=2.75cm]{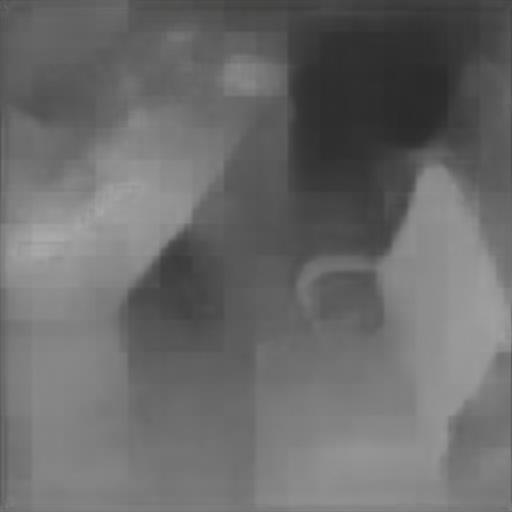}
		\captionsetup{labelformat=empty}
		\captionsetup{justification=centering}
		\caption*{(b)}
	\end{minipage}
	\begin{minipage}{.16\textwidth}
		\centering
	\caption*{\emph{SSIM:0.9201}}
	\vskip -12pt
	\includegraphics[height=2.75cm]{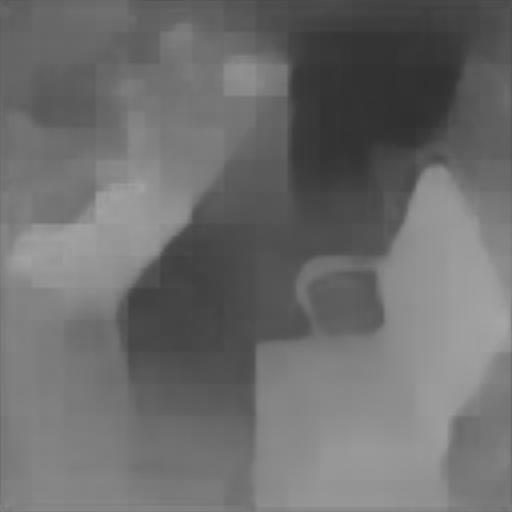}
		\captionsetup{labelformat=empty}
		\captionsetup{justification=centering}
		\caption*{(c)}
	\end{minipage} 
	\begin{minipage}{.16\textwidth}
		\centering
	\caption*{\emph{SSIM:0.9213}}
	\vskip -12pt
	\includegraphics[height=2.75cm]{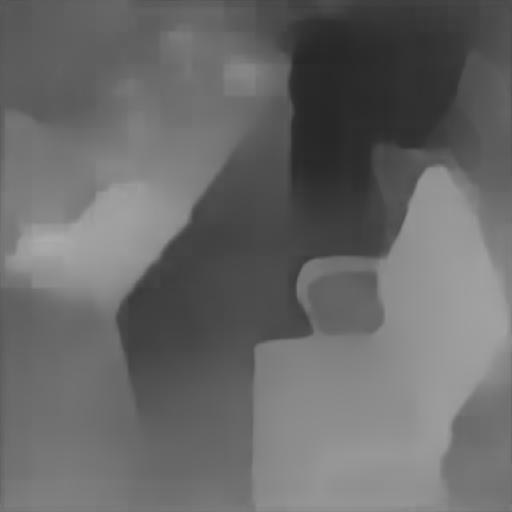}
		\captionsetup{labelformat=empty}
		\captionsetup{justification=centering}
		\caption*{(d)}
	\end{minipage}
	\begin{minipage}{.16\textwidth}
		\centering
	\caption*{\emph{SSIM:\textbf{0.9283}}}
	\vskip -12pt
	\includegraphics[height=2.75cm]{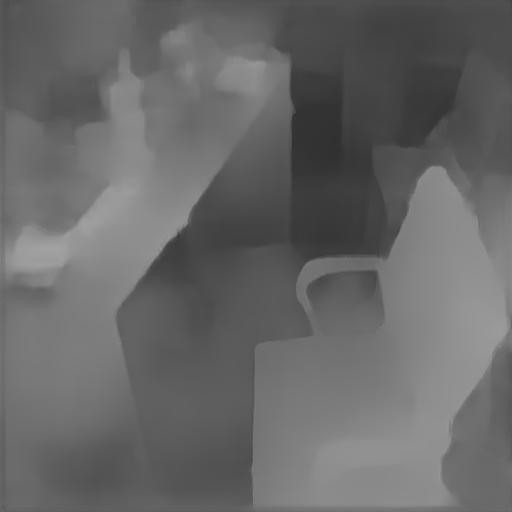}
		\captionsetup{labelformat=empty}
		\captionsetup{justification=centering}
		\caption*{(e)}
	\end{minipage}
	\begin{minipage}{.16\textwidth}
		\centering
	\caption*{\emph{SSIM:1}}
	\vskip -12pt
	\includegraphics[height=2.75cm]{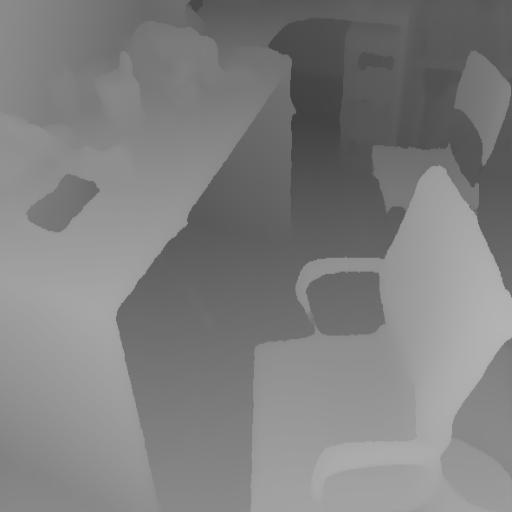}\\
		\captionsetup{labelformat=empty}
		\captionsetup{justification=centering}
		\caption*{(f)}
	\end{minipage}
	\vskip -15pt\caption{Transmission map estimation results using different modules. (a) {DED}; (b). {DED-MLP}; (c).{DED-MLP-GRA};  (d).  {DED-MLP-EP}; (e). DCPDN; (f) Target. It can be observed that the multi-level pooling module is able to refine better global structure of objects in the image (observed from (a) and (b) ), the edge-preserving loss can preserve much sharper edges (comparing (b), (c) and (d)) and  the final joint-discriminator can better refine the detail for small objects (comparing (d) and (e)).  }
	\label{baseline}
	\vskip -8pt
\end{figure*}

Based  on these observations and inspired by the gradient loss used in depth estimation \cite{depth_gra_zhou,depth_gra} as well as the use of perceptual loss in low-level vision tasks \cite{perceptual_loss,he_2018_CVPR_derain},  we propose a new edge-preserving loss function that is composed of three different parts: $L2$ loss, two-directional gradient loss, and feature edge loss, defined as follows
\begin{equation}
\label{eq:loss_trans}
L^E =  \lambda_{E,l_2} L_{E,l_2} + \lambda_{E,g} L_{E,g} + \lambda_{E,f} L_{E,f},
\end{equation}
where $L^E$ indicates the overall edge-preserving loss, $L_{E,l_2}$ indicates the L2 loss, $L_{E,g}$ indicates the two-directional (horizontal and vertical) gradient loss and $L_{E,f}$ is the feature loss. $L_{E,g}$ is defined as follows
 \begin{equation}
 \label{eq:gra}
 \begin{split}
 L_{E,g}=&\sum_{w,h}^{}\|(H_x({G_t({I}})))_{w,h}-(H_x(t))_{w,h}\|_2\\
 &+\|(H_y({G_t({I}})))_{w,h}-(H_y(t))_{w,h}\|_2,
 \end{split}
 \end{equation}
where $H_x$ and $H_y$ are operators that compute image gradients along rows (horizontal) and columns (vertical), respectively and  $w\times h$ indicates  the width and height  of the output feature map.  The feature loss is defined as
\begin{equation}
\label{eq:dehaze_loss_perc}
 \begin{split}
  L_{E,f}=& \sum_{c_1,w_1,h_1}^{} \|(V_1(G_t({{I}})))_{c_1,w_1,h_1}-(V_1
 ({t}))_{c_1,w_1,h_1}\|_2 \\
 +&\sum_{c_2,w_2,h_2}^{} \|(V_2(G_t({{I}})))_{c_2,w_2,h_2}-(V_2
  ({t}))_{c_2,w_2,h_2}\|_2,
  \end{split}
\end{equation}
 where $V_i$ represents a CNN structure and $c_i,w_i,h_i$ are the dimensions of the corresponding low-level feature in $V_i$. As the edge information is preserved in the low-level features, we adopt the layers before relu1-1 and relu2-1 of VGG-16 \cite{vgg} as the edge extractors $V_1$ and $V_2$, respectively.  Here, $ \lambda_{E,l_2}, \lambda_{E,g},$ and $\lambda_{E,f}$ are weights to balance the loss function.
 
 \begin{figure*}[t]
 	\centering
	\vskip-6pt
 	\begin{minipage}{.12\textwidth}
 		\centering
 	\caption*{\emph{SSIM: 0.7654}}
 	\vskip -12pt
 	\includegraphics[width=1\textwidth]{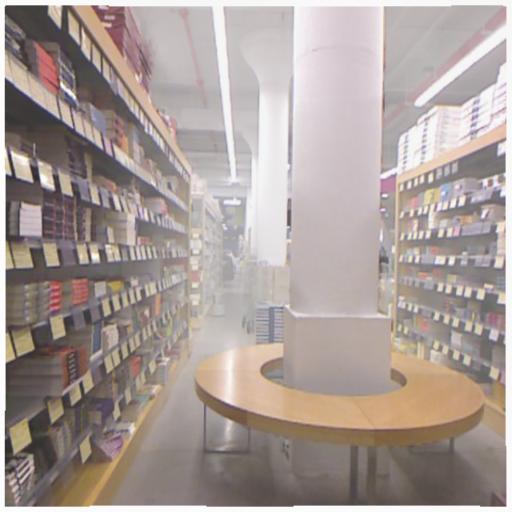}
 		\captionsetup{labelformat=empty}
 		\captionsetup{justification=centering}
 	\end{minipage} 
 	\begin{minipage}{.12\textwidth}
 		\centering
 	\caption*{\emph{SSIM: 0.9382}}
 	\vskip -12pt
 	\includegraphics[width=1\textwidth]{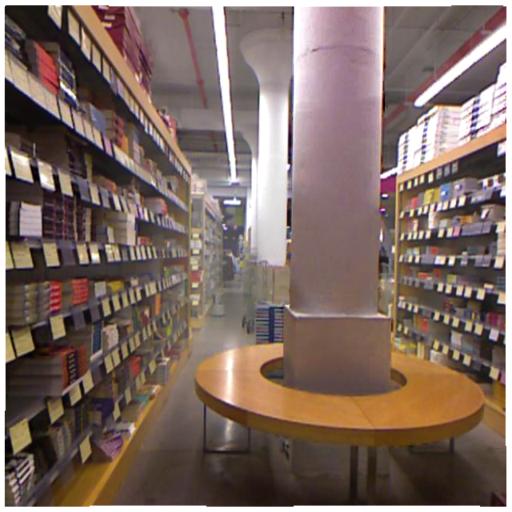}
 		\captionsetup{labelformat=empty}
 		\captionsetup{justification=centering}
 	\end{minipage}  
 	\begin{minipage}{.12\textwidth}
 		\centering
 	\caption*{\emph{SSIM: 0.8637}}
 	\vskip -12pt
 	\includegraphics[width=1\textwidth]{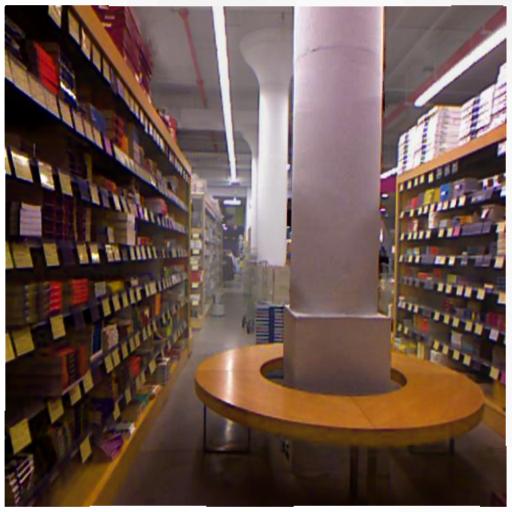}
 		\captionsetup{labelformat=empty}
 		\captionsetup{justification=centering}
 	\end{minipage} 
 	\begin{minipage}{.12\textwidth}
 		\centering
 	\caption*{\emph{SSIM: 0.9005}}
 	\vskip -12pt
 	\includegraphics[width=1\textwidth]{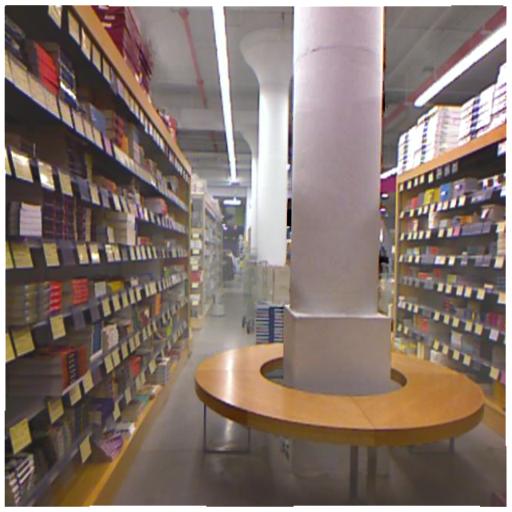}
 		\captionsetup{labelformat=empty}
 		\captionsetup{justification=centering}
 	\end{minipage} 
 	\begin{minipage}{.12\textwidth}
 		\centering
 	\caption*{\emph{SSIM: 0.8683}}
 	\vskip -12pt
 	\includegraphics[width=1\textwidth]{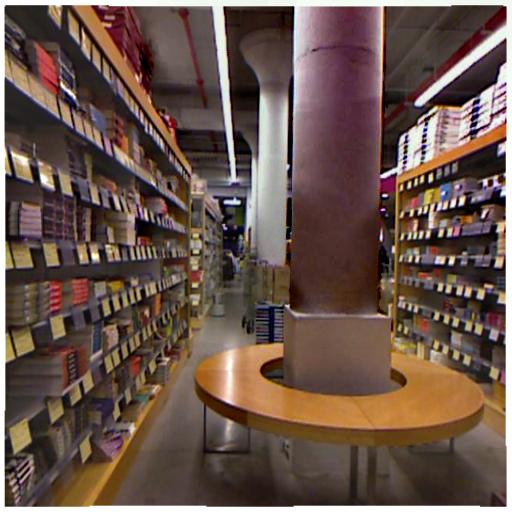}
 		\captionsetup{labelformat=empty}
 		\captionsetup{justification=centering}
 	\end{minipage} 
 	\begin{minipage}{.12\textwidth}
 		\centering
 	\caption*{\emph{SSIM: {0.9200}}}
 	\vskip -12pt
 	\includegraphics[width=1\textwidth]{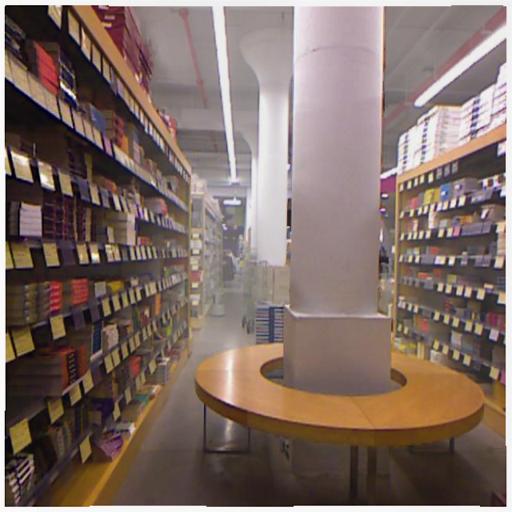}
 		\captionsetup{labelformat=empty}
 		\captionsetup{justification=centering}
 	\end{minipage}
 	\begin{minipage}{.12\textwidth}
 		\centering
 	\caption*{\emph{SSIM: \textbf{0.9777}}}
 	\vskip -12pt
 	\includegraphics[width=1\textwidth]{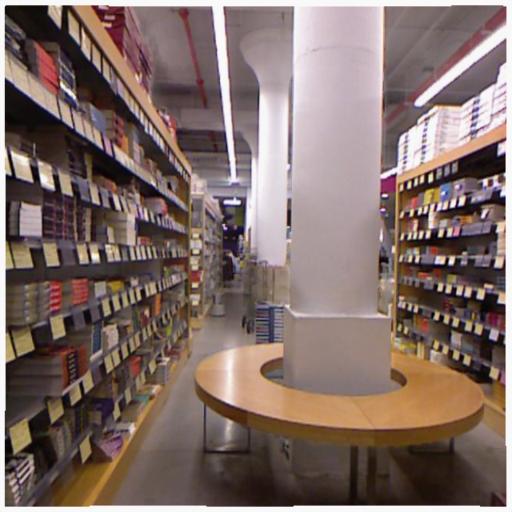}
 		\captionsetup{labelformat=empty}
 		\captionsetup{justification=centering}
 	\end{minipage} 
 	\begin{minipage}{.12\textwidth}
 		\centering
 	\caption*{\emph{SSIM:1}}
 	\vskip -12pt
 	\includegraphics[width=1\textwidth]{syn_result_image//399_gt.jpg}
 		\captionsetup{labelformat=empty}
 		\captionsetup{justification=centering}
 	\end{minipage}\\\vskip+6pt
 	\begin{minipage}{.12\textwidth}
 		\centering
	 	\caption*{\emph{SSIM: 0.6642}}
 	\vskip -12pt
 	\includegraphics[width=1\textwidth]{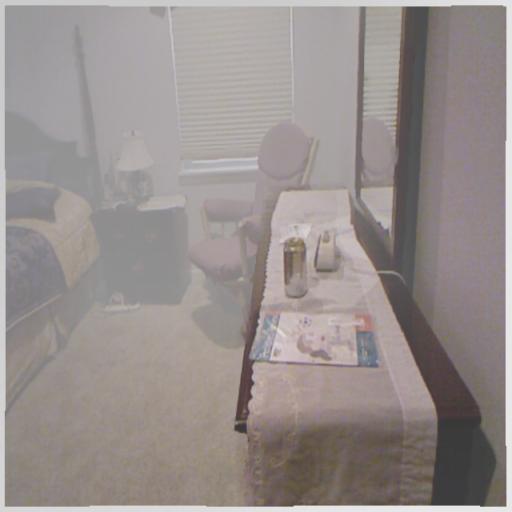}
 		\captionsetup{labelformat=empty}
 		\captionsetup{justification=centering}
 		\caption*{Input\\\qquad}
 	\end{minipage} 
 	\begin{minipage}{.12\textwidth}
 		\centering
	 	\caption*{\emph{SSIM: 0.8371}}
 	\vskip -12pt
 	\includegraphics[width=1\textwidth]{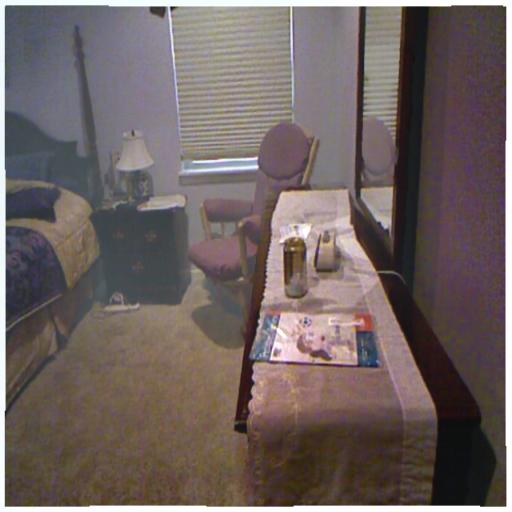}
 		\captionsetup{labelformat=empty}
 		\captionsetup{justification=centering}
 		\caption*{He \emph{et al.}  \cite{dehaz_2009_dark} \\\qquad}
 	\end{minipage}  
 	\begin{minipage}{.12\textwidth}
 		\centering
	 	\caption*{\emph{SSIM: 0.8117}}
 	\vskip -12pt
 	\includegraphics[width=1\textwidth]{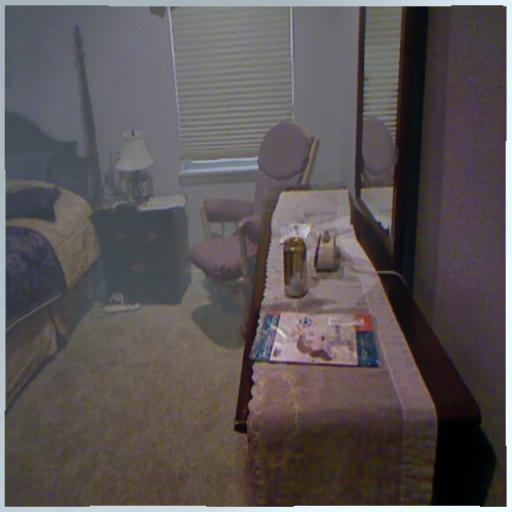}
 		\captionsetup{labelformat=empty}
 		\captionsetup{justification=centering}
 		\caption*{Zhu \emph{et al.} \cite{dehaze_fast15}\\\qquad}
 	\end{minipage} 
 	\begin{minipage}{.12\textwidth}
 		\centering
	 	\caption*{\emph{SSIM: 0.8364}}
 	\vskip -12pt
 	\includegraphics[width=1\textwidth]{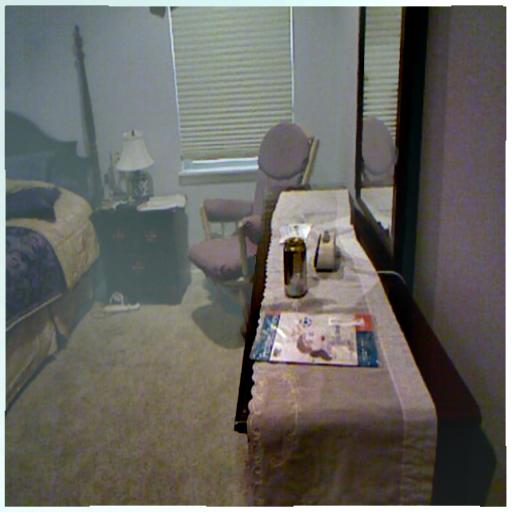}
 		\captionsetup{labelformat=empty}
 		\captionsetup{justification=centering}
 		\caption*{Ren \emph{et al.}  \cite{dehaze_2016_eccv} \\\qquad}
 	\end{minipage} 
 	\begin{minipage}{.12\textwidth}
 		\centering
	 	\caption*{\emph{SSIM: 0.8575}}
 	\vskip -12pt
 	\includegraphics[width=1\textwidth]{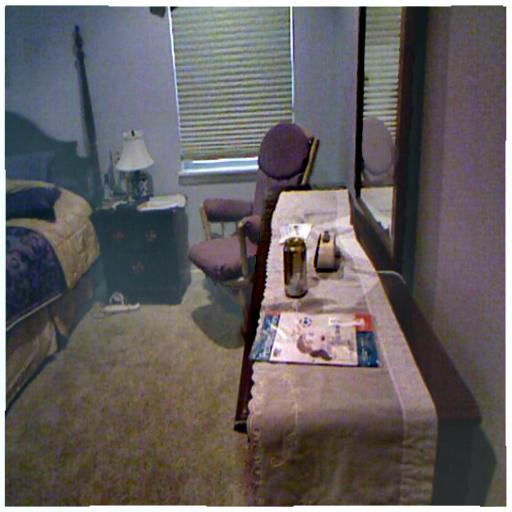}
 		\captionsetup{labelformat=empty}
 		\captionsetup{justification=centering}
 		\caption*{Berman \emph{et al.} \cite{dehaze_2016_dana,dehaze_iccp217}}
 	\end{minipage} 
 	\begin{minipage}{.12\textwidth}
 		\centering
	 	\caption*{\emph{SSIM: 0.7691}}
 	\vskip -12pt
 	\includegraphics[width=1\textwidth]{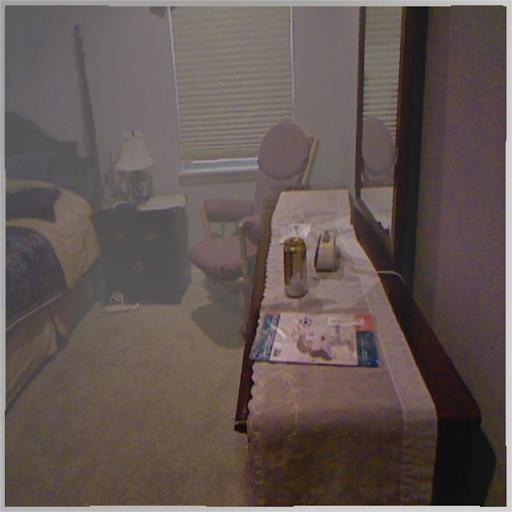}
 		\captionsetup{labelformat=empty}
 		\captionsetup{justification=centering}
 		\caption*{Li \emph{et al.} \cite{dehaze_iccv17} \\\qquad}
 	\end{minipage}
 	\begin{minipage}{.12\textwidth}
 		\centering
	 	\caption*{\textbf{\emph{SSIM: 0.9325}}}
 	\vskip -12pt
 	\includegraphics[width=1\textwidth]{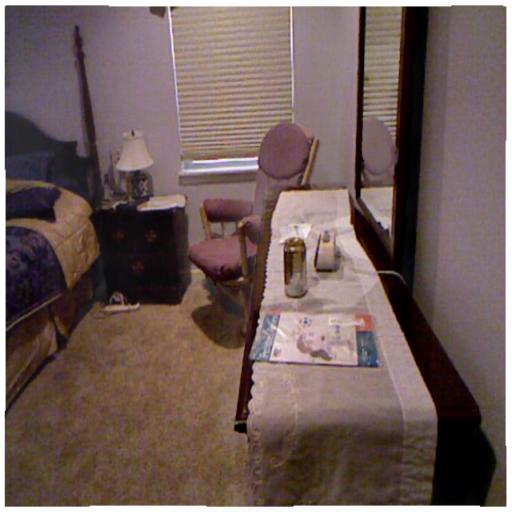}
 		\captionsetup{labelformat=empty}
 		\captionsetup{justification=centering}
 		\caption*{DCPDN \\\qquad}
 	\end{minipage}
 	\begin{minipage}{.12\textwidth}
 		\centering
 	\caption*{\emph{SSIM: 1}}
 	\vskip -12pt
 	\includegraphics[width=1\textwidth]{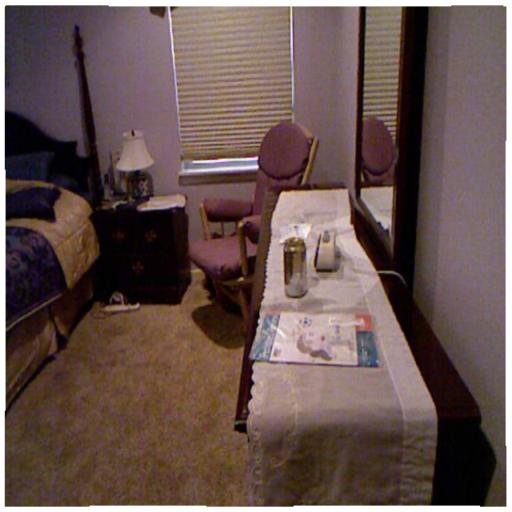}
 		\captionsetup{labelformat=empty}
 		\captionsetup{justification=centering}
 		\caption*{GT\\\qquad}
 	\end{minipage}
 	\vskip -17pt\caption{Dehazing results from the synthetic test datasets \textbf{TestA} (first row) and \textbf{TestB} (second row).}\label{fig:syn_test}
 	\label{fig:synthetic}
\vskip -9pt
 \end{figure*}

\subsection{Overall Loss Function} 
The proposed DCPDN architecture is trained using the following four loss functions
\begin{equation}
\label{eq:loss_all}
L = L^t + L^a + L^d + \lambda_j L^j,
\end{equation}
where $L^t$ is composed of the edge-preserving loss $L^E$, $L^a$ is composed of the traditional $L2$ loss in predicting the atmospheric light and $L^d$ represents the dehazing loss, which is also composed of the $L2$ loss only. $L^j$, which is denoted as the joint discriminator loss \footnote{To address the vanishing gradients problem for the generator, we also minimize \eqref{eq:trans_loss_adv} rather  than the first two rows in \eqref{eq:GAN_jd} \cite{gan,gan_tutorial}.}, is defined as follows
\begin{equation}
 \label{eq:trans_loss_adv}
   L^j= -\log({D_{joint}}({G_t}(I))-\log({D_{joint}}({G_d}(I)).
\end{equation}
Here $\lambda_j$ is a constant. 

\subsection{Stage-wise Learning}
During experiments, we found that directly training the whole network from scratch with the complex loss Eq.~\ref{eq:loss_all} is difficult and the network converges very slowly.  A possible reason may be due to the gradient diffusion caused by different tasks.  For example, gradients from the de-hazed image loss may `distract'  the gradients from the loss of the transmission map initially, resulting in the slower convergence.  To address this issue and to speed up the training, a stage-wise learning strategy is introduced, which has been used in different applications such as multi-model recognition \cite{stage_wise_recog} and feature learning \cite{stage_wise_feature}.  Hence, the information in the training data is presented to the network gradually. In other words,  different tasks are learned progressively.  Firstly, we optimize each task separately by not updating the other task simultaneously.  After the `initialization' for each task, we fine-tune the whole network all together by optimizing all three tasks jointly.   

\section{Experimental Results}
In this section, we demonstrate the effectiveness of the proposed approach by conducting various experiments on two synthetic datasets and a real-world dataset. All the results are compared with five state-of-the-art methods: He \emph{et al.} (CVPR'09) \cite{dehaz_2009_dark}, Zhu  \emph{et al} (TIP'15) \cite{dehaze_fast15}, Ren \emph{et al.} \cite{dehaze_2016_eccv} (ECCV'16), Berman \emph{et al.} \cite{dehaze_2016_dana,dehaze_iccp217} (CVPR'16 and ICCP'17) and Li \emph{et al.} \cite{dehaze_iccv17} (ICCV'17). In addition, we conduct an ablation study to demonstrate the effectiveness of each module of our network.

\subsection{Datasets}
Similar to the existing deep learning-based dehazing methods \cite{dehaze_2016_eccv,dehaze_2016_tip,dehaze_iccv17,dehaze_2017_joint}, we synthesize the training samples \emph{\{Hazy /Clean /Transmission Map /Atmosphere Light\}} based on \eqref{eq: intro}. During  synthesis, four atmospheric light conditions $A \in[0.5,1]$ and the scattering coefficient $\beta \in[0.4,1.6]$ are randomly sampled to generate their corresponding hazy images, transmission maps and atmospheric light maps. A random set of 1000 images are selected from the NYU-depth2 dataset \cite{nyudepth} to generate the training set. Hence, there are in total 4000 training images, denoted as \textbf{TrainA}. Similarly, a test dataset \textbf{TestA} consisting of 400 (100$\times$4) images also from the NYU-depth2 are obtained. We ensure that none of the testing images are in the training set. To demonstrate the generalization ability of our network to other datasets, we synthesize 200 \emph{\{Hazy /Clean /Transmission Map /Atmosphere Light\}} images from both the Middlebury stereo database (40) \cite{depth_middlebbury} and also the Sun3D dataset  (160) \cite{sunny_3d_depth} as the \textbf{TestB} set.


\begin{table}[]
	\centering
	\caption{Quantitative SSIM results for ablation study evaluated on synthetic \textbf{TestA} and \textbf{TestB} datasets. }
    \resizebox{0.45\textwidth}{!}{%
\begin{tabular}{cccccc}
 &  &  & TestA &  &  \\ \hline
\multicolumn{1}{|c|}{} & \multicolumn{1}{c|}{DED} & \multicolumn{1}{c|}{DED-MLP} & \multicolumn{1}{c|}{DED-MLP-GRA} & \multicolumn{1}{c|}{DED-MLP-EP} & \multicolumn{1}{c|}{DCPDN} \\ \hline
\multicolumn{1}{|c|}{Transmission} & \multicolumn{1}{c|}{0.9555} & \multicolumn{1}{c|}{0.9652} & \multicolumn{1}{c|}{0.9687} & \multicolumn{1}{c|}{0.9732} & \multicolumn{1}{c|}{0.9776} \\ \hline
\multicolumn{1}{|c|}{Image} & \multicolumn{1}{c|}{0.9252} & \multicolumn{1}{c|}{0.9402} & \multicolumn{1}{c|}{0.9489} & \multicolumn{1}{c|}{0.9530} & \multicolumn{1}{c|}{0.9560} \\ \hline
 &  &  & TestB &  &  \\ \hline
\multicolumn{1}{|c|}{Transmission} & \multicolumn{1}{c|}{0.9033} & \multicolumn{1}{c|}{0.9109} & \multicolumn{1}{c|}{0.9239} & \multicolumn{1}{c|}{0.9276} & \multicolumn{1}{c|}{0.9352} \\ \hline
\multicolumn{1}{|c|}{Image} & \multicolumn{1}{c|}{0.8474} & \multicolumn{1}{c|}{0.8503} & \multicolumn{1}{c|}{0.8582} & \multicolumn{1}{c|}{0.8652} & \multicolumn{1}{c|}{0.8746} \\ \hline
\end{tabular}}
	\label{ta:base_quan1}
\end{table}

\begin{figure*}[t]
	\centering
	\vskip-6pt
	\begin{minipage}{.135\textwidth}
		\centering
		\includegraphics[width=2.4cm, height=2.1cm]{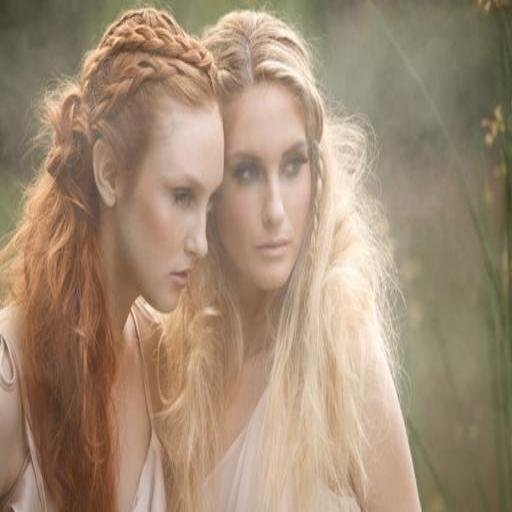}
		\captionsetup{labelformat=empty}
		\captionsetup{justification=centering}
	\end{minipage} 
	\begin{minipage}{.135\textwidth}
		\centering
		\includegraphics[width=2.4cm, height=2.1cm]{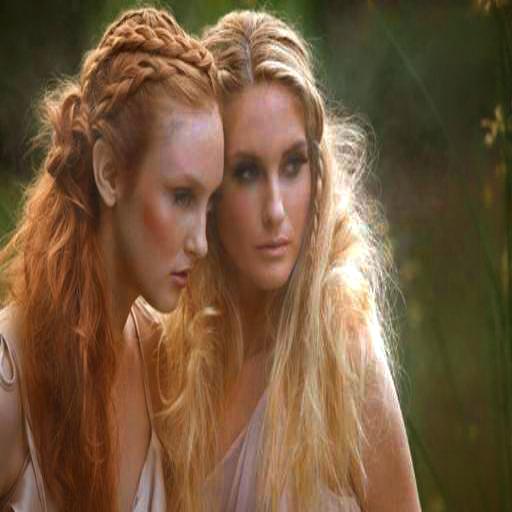}
		\captionsetup{labelformat=empty}
		\captionsetup{justification=centering}
	\end{minipage} 
	\begin{minipage}{.135\textwidth}
		\centering
		\includegraphics[width=2.4cm, height=2.1cm]{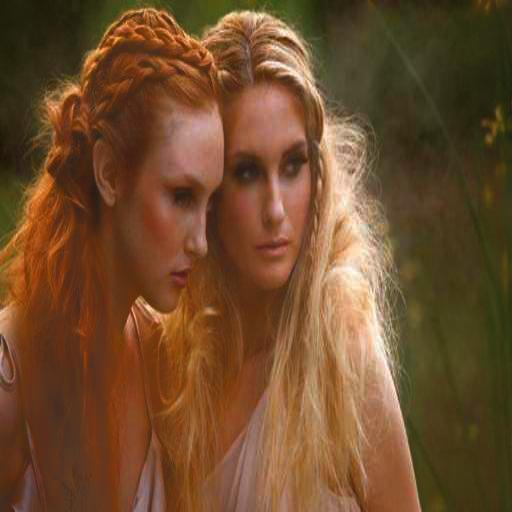}
		\captionsetup{labelformat=empty}
		\captionsetup{justification=centering}
	\end{minipage} 
	\begin{minipage}{.135\textwidth}
		\centering
		\includegraphics[width=2.4cm, height=2.1cm]{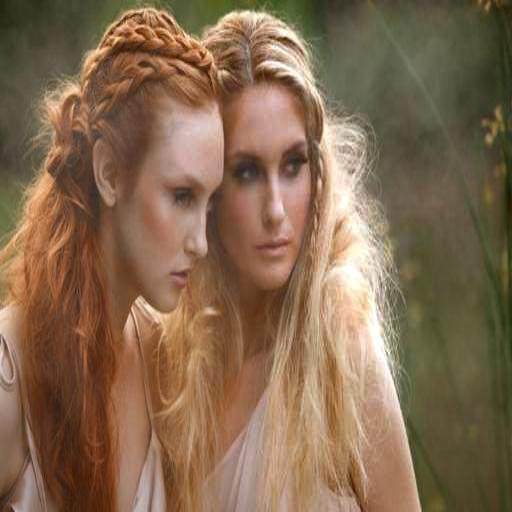}
		\captionsetup{labelformat=empty}
		\captionsetup{justification=centering}
	\end{minipage} 
	\begin{minipage}{.135\textwidth}
		\centering
		\includegraphics[width=2.4cm, height=2.1cm]{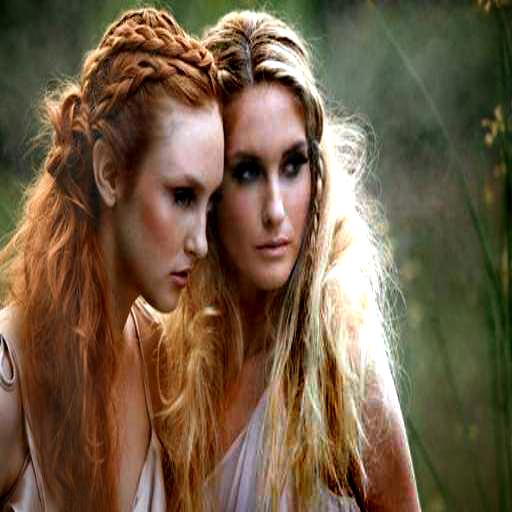}
		\captionsetup{labelformat=empty}
		\captionsetup{justification=centering}
	\end{minipage} 
	\begin{minipage}{.135\textwidth}
		\centering
		\includegraphics[width=2.4cm, height=2.1cm]{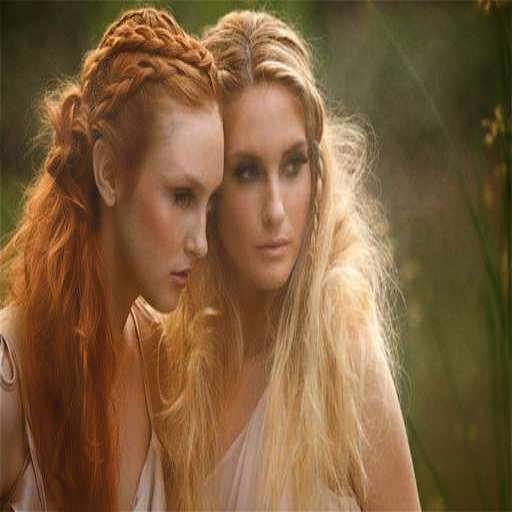}
		\captionsetup{labelformat=empty}
		\captionsetup{justification=centering}
	\end{minipage}
	\begin{minipage}{.135\textwidth}
		\centering
		\includegraphics[width=2.4cm, height=2.1cm]{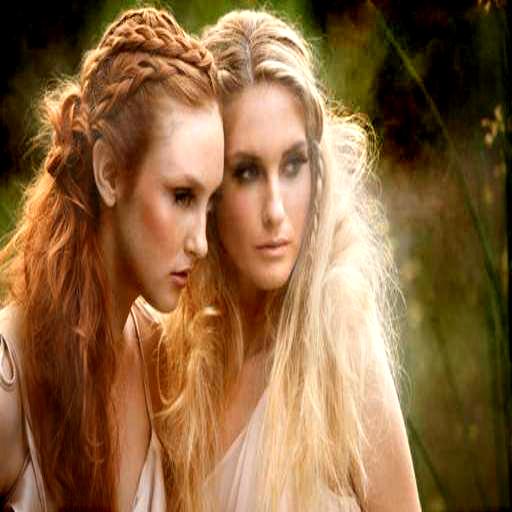}
		\captionsetup{labelformat=empty}
		\captionsetup{justification=centering}
	\end{minipage} \\		\vskip+4pt
	\begin{minipage}{.135\textwidth}
		\centering
		\includegraphics[width=2.4cm, height=2.1cm]{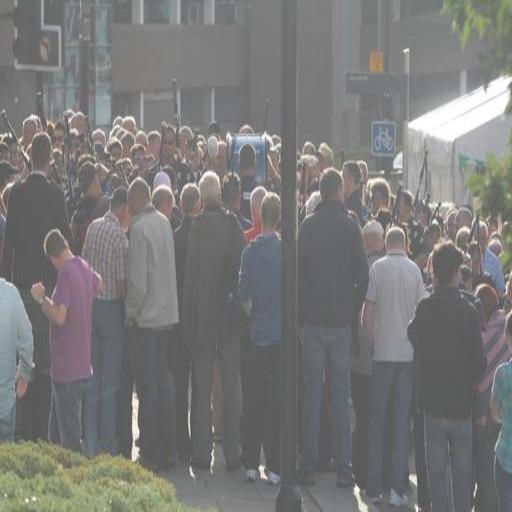}
		\captionsetup{labelformat=empty}
		\captionsetup{justification=centering}
	\end{minipage} 
	\begin{minipage}{.135\textwidth}
		\centering
		\includegraphics[width=2.4cm, height=2.1cm]{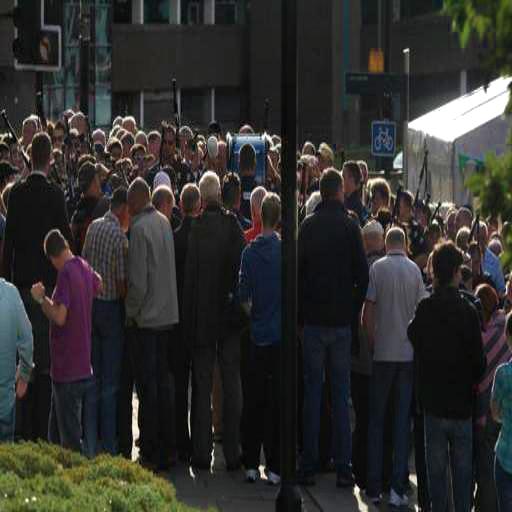}
		\captionsetup{labelformat=empty}
		\captionsetup{justification=centering}
	\end{minipage} 
	\begin{minipage}{.135\textwidth}
		\centering
		\includegraphics[width=2.4cm, height=2.1cm]{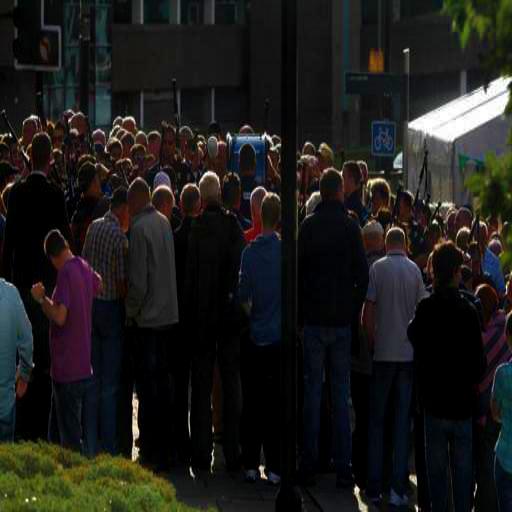}
		\captionsetup{labelformat=empty}
		\captionsetup{justification=centering}
	\end{minipage} 
	\begin{minipage}{.135\textwidth}
		\centering
		\includegraphics[width=2.4cm, height=2.1cm]{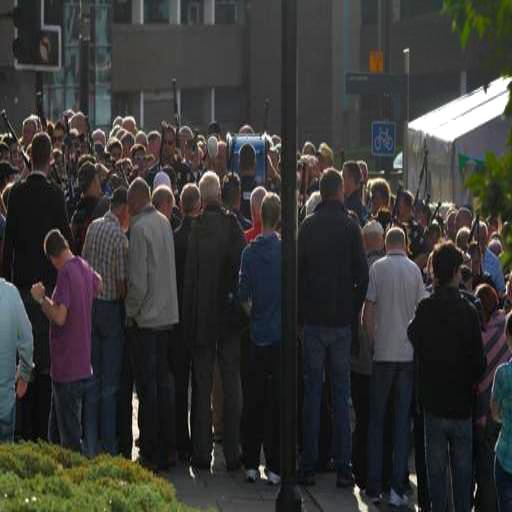}
		\captionsetup{labelformat=empty}
		\captionsetup{justification=centering}
	\end{minipage} 
	\begin{minipage}{.135\textwidth}
		\centering
		\includegraphics[width=2.4cm, height=2.1cm]{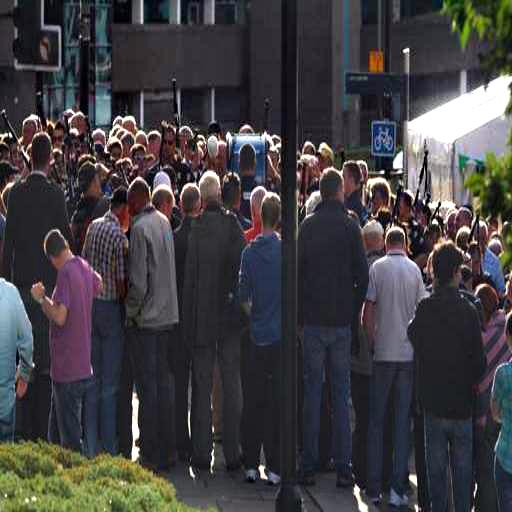}
		\captionsetup{labelformat=empty}
		\captionsetup{justification=centering}
	\end{minipage} 
	\begin{minipage}{.135\textwidth}
		\centering
		\includegraphics[width=2.4cm, height=2.1cm]{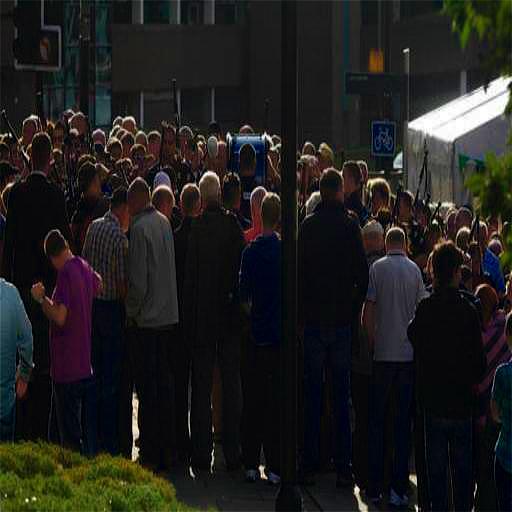}
		\captionsetup{labelformat=empty}
		\captionsetup{justification=centering}
	\end{minipage}
	\begin{minipage}{.135\textwidth}
		\centering
		\includegraphics[width=2.4cm, height=2.1cm]{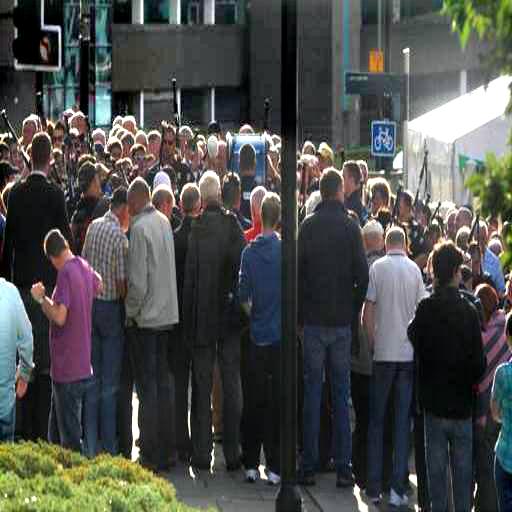}
		\captionsetup{labelformat=empty}
		\captionsetup{justification=centering}
	\end{minipage} \\		\vskip+4pt
	\begin{minipage}{.135\textwidth}
		\centering
		\includegraphics[width=2.4cm, height=1.5cm]{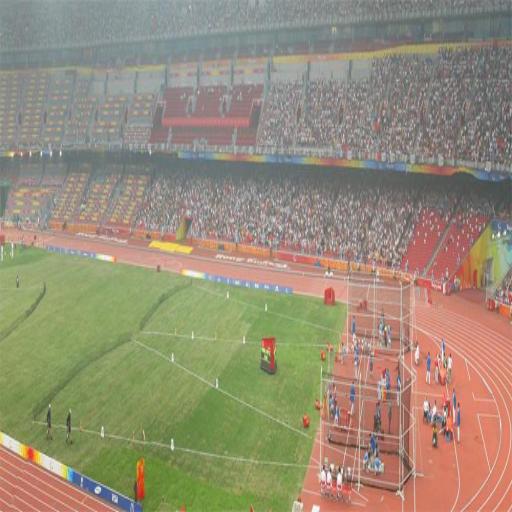}
		\captionsetup{labelformat=empty}
		\captionsetup{justification=centering}
	\end{minipage} 
	\begin{minipage}{.135\textwidth}
		\centering
		\includegraphics[width=2.4cm, height=1.5cm]{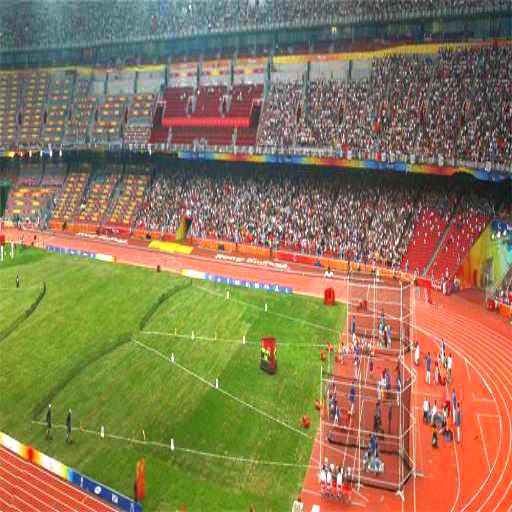}
		\captionsetup{labelformat=empty}
		\captionsetup{justification=centering}
	\end{minipage} 
	\begin{minipage}{.135\textwidth}
		\centering
		\includegraphics[width=2.4cm, height=1.5cm]{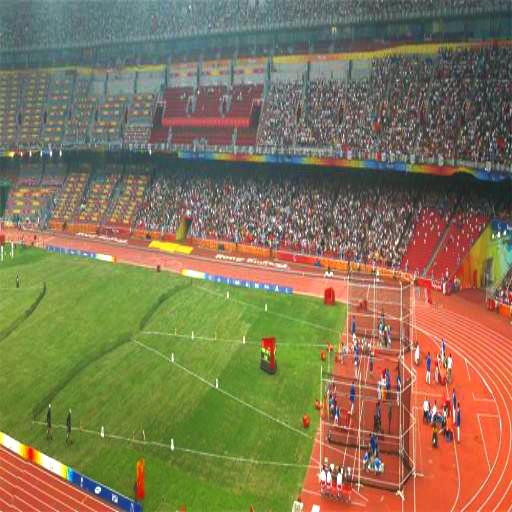}
		\captionsetup{labelformat=empty}
		\captionsetup{justification=centering}
	\end{minipage} 
	\begin{minipage}{.135\textwidth}
		\centering
		\includegraphics[width=2.4cm, height=1.5cm]{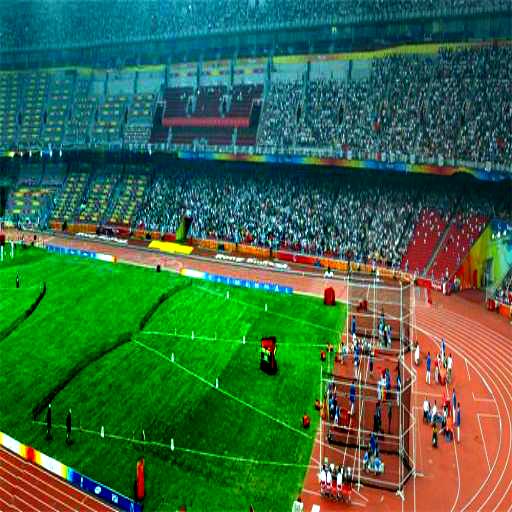}
		\captionsetup{labelformat=empty}
		\captionsetup{justification=centering}
	\end{minipage} 
	\begin{minipage}{.135\textwidth}
		\centering
		\includegraphics[width=2.4cm, height=1.5cm]{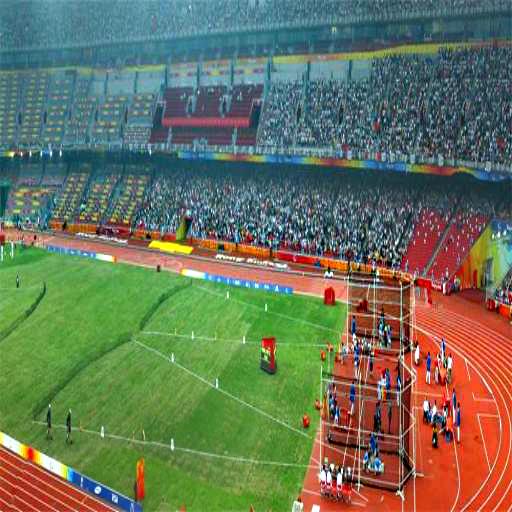}
		\captionsetup{labelformat=empty}
		\captionsetup{justification=centering}
	\end{minipage} 
	\begin{minipage}{.135\textwidth}
		\centering
		\includegraphics[width=2.4cm, height=1.5cm]{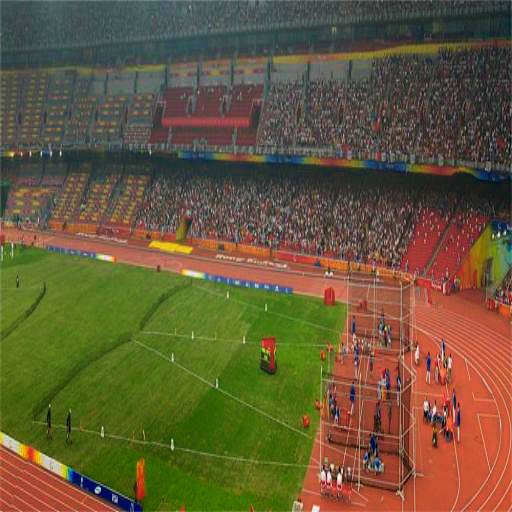}
		\captionsetup{labelformat=empty}
		\captionsetup{justification=centering}
	\end{minipage}
	\begin{minipage}{.135\textwidth}
		\centering
		\includegraphics[width=2.4cm, height=1.5cm]{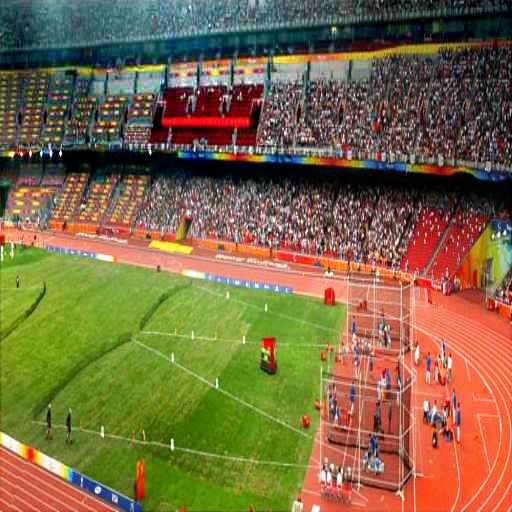}
		\captionsetup{labelformat=empty}
		\captionsetup{justification=centering}
	\end{minipage} \\		\vskip+4pt
	\begin{minipage}{.135\textwidth}
		\centering
		\includegraphics[width=2.4cm, height=2.9cm]{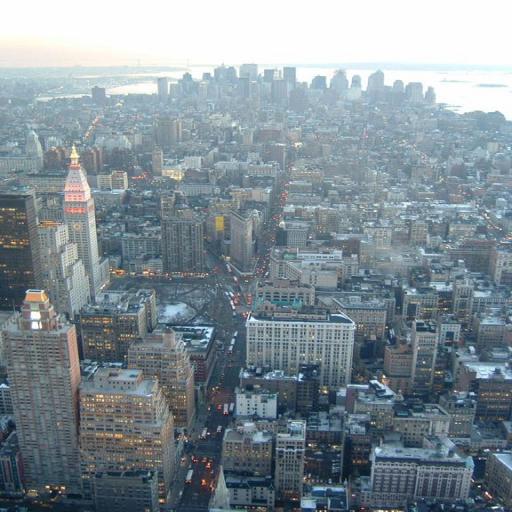}
		\captionsetup{labelformat=empty}
		\captionsetup{justification=centering}
		\caption*{Input\\ \qquad}
	\end{minipage} 
	\begin{minipage}{.135\textwidth}
		\centering
		\includegraphics[width=2.4cm, height=2.9cm]{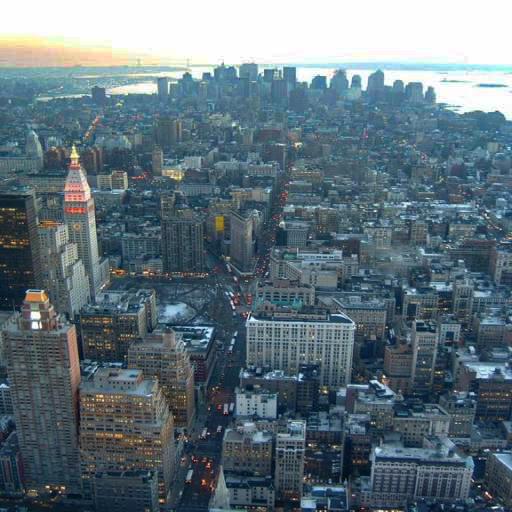}
		\captionsetup{labelformat=empty}
		\captionsetup{justification=centering}
		\caption*{He. \emph{et al.} (CVPR'09) \cite{dehaz_2009_dark}}
	\end{minipage} 
	\begin{minipage}{.135\textwidth}
		\centering
		\includegraphics[width=2.4cm, height=2.9cm]{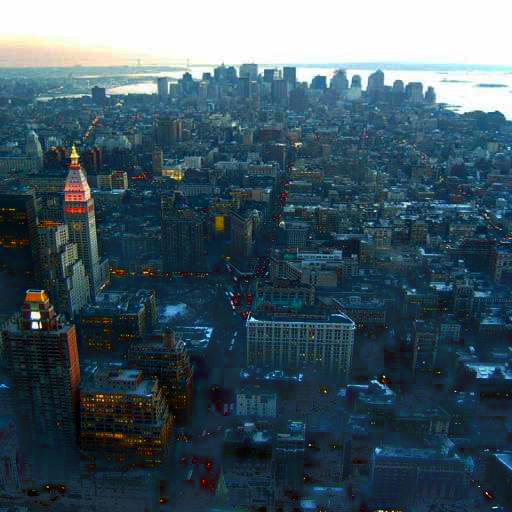}
		\captionsetup{labelformat=empty}
		\captionsetup{justification=centering}
		\caption*{Zhu. \emph{et al.} (TIP'15) \cite{dehaze_fast15}}
	\end{minipage} 
	\begin{minipage}{.135\textwidth}
		\centering
		\includegraphics[width=2.4cm, height=2.9cm]{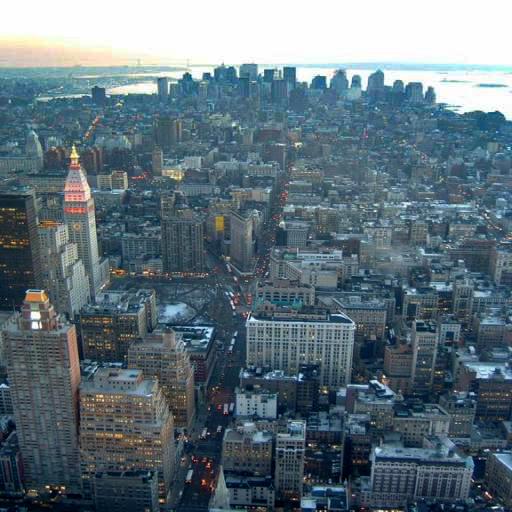}
		\captionsetup{labelformat=empty}
		\captionsetup{justification=centering}
		\caption*{Ren. \emph{et al.} (ECCV'16)\cite{dehaze_2016_eccv}}
	\end{minipage}
	\begin{minipage}{.135\textwidth}
		\centering
		\includegraphics[width=2.4cm, height=2.9cm]{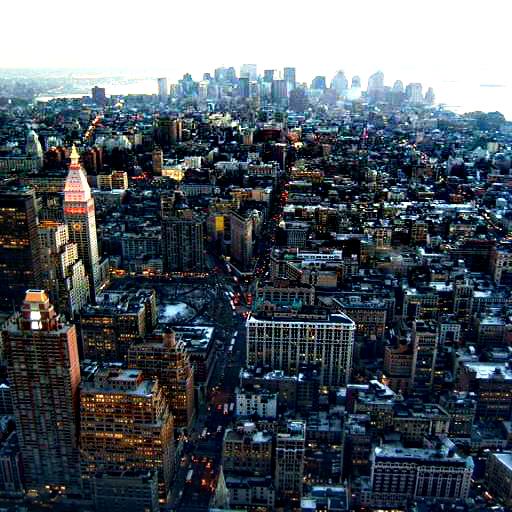}
		\captionsetup{labelformat=empty}
		\captionsetup{justification=centering}
		\caption*{Berman. \emph{et al.} (CVPR'16) \cite{dehaze_2016_dana,dehaze_iccp217}}
	\end{minipage} 
	\begin{minipage}{.135\textwidth}
		\centering
		\includegraphics[width=2.4cm, height=2.9cm]{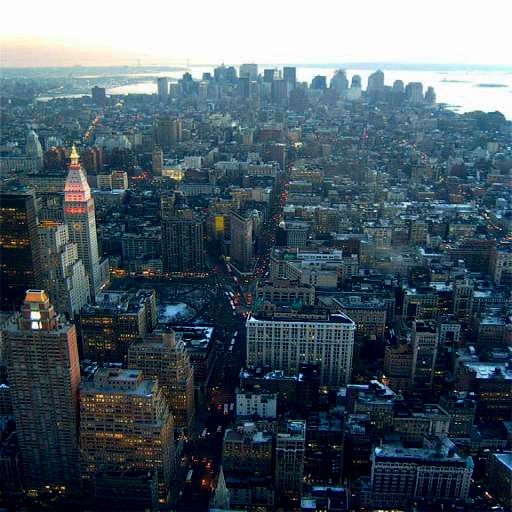}
		\captionsetup{labelformat=empty}
		\captionsetup{justification=centering}
		\caption*{Li. \emph{et al.} (ICCV'17) \cite{dehaze_iccv17} }
	\end{minipage}
	\begin{minipage}{.135\textwidth}
		\centering
		\includegraphics[width=2.4cm, height=2.9cm]{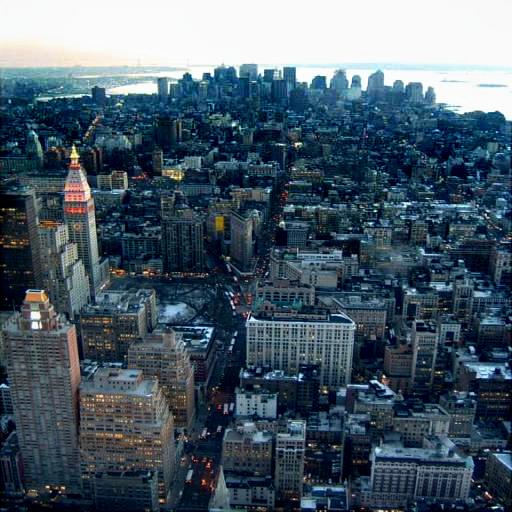}
		\captionsetup{labelformat=empty}
		\captionsetup{justification=centering}
		\caption*{DCPDN\\ \qquad}
	\end{minipage}	
	\vskip -8pt\caption{Dehazing results evaluated on real-world images released  by the authors of previous methods. }\label{fig:nature_old}
\vskip -8pt
\end{figure*}

\subsection{Training Details}
We choose $\lambda_{E,l_2}=1$, $\lambda_{E,g}=0.5$, $\lambda_{E,f}=0.8$ for the loss in estimating the transmission map and $\lambda_j=0.25$ for optimizing the joint discriminator.  During training, we use ADAM  as the optimization algorithm with learning rate  of $2\times 10^{-3}$ for both generator and discriminator and batch size of  1.   All the training samples are resized to $512\times 512$. We trained the network for 400000 iterations.  All the parameters are chosen via cross-validation. 
\begin{table}[t]
\vskip -4pt
	\centering
    \caption{Quantitative SSIM results on the synthetic \textbf{TestA} dataset. }
    \resizebox{0.45\textwidth}{!}{%
\begin{tabular}{|c|c|c|c|c|c|c|c|}
	\hline
	& Input &  \begin{tabular}[c]{@{}l@{}}He. \emph{et al.} \cite{dehaz_2009_dark}\\ (CVPR'09)\end{tabular} & \begin{tabular}[c]{@{}l@{}}Zhu. \emph{et al.} \cite{dehaze_fast15}\\ (TIP'15) \end{tabular}   & \begin{tabular}[c]{@{}l@{}} Ren. \emph{et al.} \cite{dehaze_2016_eccv} \\ (ECCV'16) \end{tabular} & \begin{tabular}[c]{@{}l@{}}Berman. \emph{et al.} \cite{dehaze_2016_dana,dehaze_iccp217} \\ (CVPR'16) \end{tabular} & \begin{tabular}[c]{@{}l@{}}Li. \emph{et al.} \cite{dehaze_iccv17} \\ (ICCV'17) \end{tabular}  & DCPDN\\ \hline\hline
	Transmission & N/A & 0.8739 & 0.8326  & N/A & 0.8675 & N/A & \textbf{0.9776} \\ \hline
	Image &  0.7041  & 0.8642   & 0.8567 & 0.8203 & 0.7959 & {0.8842} & \textbf{0.9560} \\ \hline
\end{tabular}}
	\label{ta:quantitive_depth1}
\vskip -4pt
\end{table}

\begin{table}[t]
	\centering
    \caption{Quantitative SSIM results on the synthetic \textbf{TestB} dataset. }
    \resizebox{0.45\textwidth}{!}{%
\begin{tabular}{|c|c|c|c|c|c|c|c|}
	\hline
	& Input &  \begin{tabular}[c]{@{}l@{}}He. \emph{et al.} \cite{dehaz_2009_dark}\\ (CVPR'09)\end{tabular} & \begin{tabular}[c]{@{}l@{}}Zhu. \emph{et al.} \cite{dehaze_fast15}\\ (TIP'15) \end{tabular} & \begin{tabular}[c]{@{}l@{}} Ren. \emph{et al.} \cite{dehaze_2016_eccv} \\ (ECCV'16) \end{tabular} & \begin{tabular}[c]{@{}l@{}}Berman. \emph{et al.} \cite{dehaze_2016_dana,dehaze_iccp217} \\ (CVPR'16) \end{tabular}   & \begin{tabular}[c]{@{}l@{}}Li. \emph{et al.} \cite{dehaze_iccv17} \\ (ICCV'17) \end{tabular}  & DCPDN\\ \hline\hline
	Transmission & N/A  & 0.8593 & 0.8454  & N/A & 0.8769 & N/A  & \textbf{0.9352} \\ \hline
	Image &  0.6593  & 0.7890  & 0.8253  & 0.7724 & 0.7597 & 0.8325 & \textbf{0.8746} \\ \hline
\end{tabular}}
	\label{ta:quantitive_depth2}
\vskip -4pt
\end{table}
\subsection{Ablation Study}
In order to demonstrate the improvements obtained by each  module introduced in the proposed network, we perform an ablation study involving the following five experiments: 1) Densely connected encoder decoder structure (\textbf{DED}), 2) Densely connected encoder decoder structure with multi-level pyramid pooling (\textbf{DED-MLP}), 3) Densely connected encoder decoder structure with multi-level pyramid pooling using L2 loss and gradient loss (\textbf{DED-MLP-GRA}),  4)  Densely connected encoder decoder structure with multi-level pyramid pooling using edge-preserving loss (\textbf{DED-MLP-EP}), 5) The proposed DCPDN that is composed of densely connected encoder decoder structure with multi-level pyramid pooling using edge-preserving loss and joint discriminator (\textbf{DCPDN}). \footnote{The configuration 1) \textbf{DED} and 2) \textbf{DED-MLP} are optimized  only with $L2$ loss.}

The evaluation is performed on the synthesized \textbf{TestA} and \textbf{TestB} datasets. The SSIM results averaged on both estimated transmission maps and dehazed images for the various configurations are tabulated in Table~\ref{ta:base_quan1}. Visual comparisons are shown in the Fig~\ref{baseline}.   From Fig~\ref{baseline}, we make the following observations: 1) The proposed multi-level pooling module is able to better preserve the `global' structural for objects with relatively larger scale, compared with (a) and (b). 2) The use of edge-preserving loss is able to better refine the edges in the estimated transmission map, compared with (b), (c) and (d). 3) The final joint-discriminator can further enhance the estimated transmission map by ensuring that the fine structural details are captured in the results, such as details of the small objects on the table shown in the second row in (e). The quantitative performance evaluated on both \textbf{TestA} and \textbf{TestB} also demonstrate the effectiveness of each module.

\begin{figure*}[t]
	\centering
	\begin{minipage}{.135\textwidth}
		\centering
		\includegraphics[width=2.4cm, height=2cm]{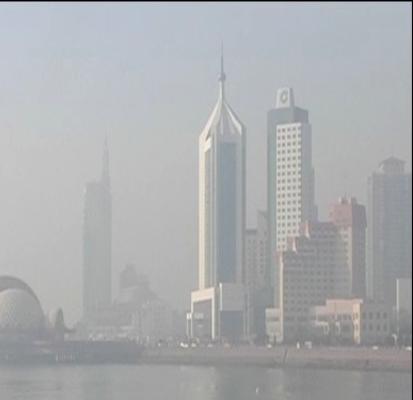}
		\captionsetup{labelformat=empty}
		\captionsetup{justification=centering}
	\end{minipage} 
	\begin{minipage}{.135\textwidth}
		\centering
		\includegraphics[width=2.4cm, height=2cm]{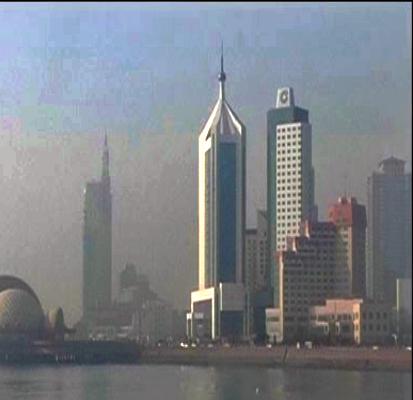}
		\captionsetup{labelformat=empty}
		\captionsetup{justification=centering}
	\end{minipage} 
	\begin{minipage}{.135\textwidth}
		\centering
		\includegraphics[width=2.4cm, height=2cm]{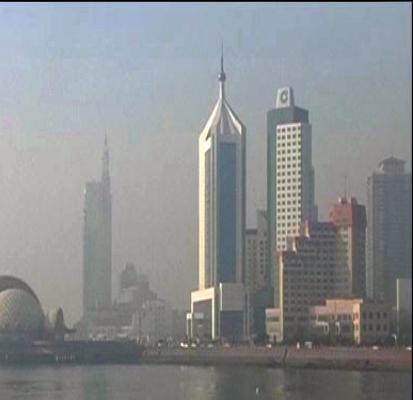}
		\captionsetup{labelformat=empty}
		\captionsetup{justification=centering}
	\end{minipage} 
	\begin{minipage}{.135\textwidth}
		\centering
		\includegraphics[width=2.4cm, height=2cm]{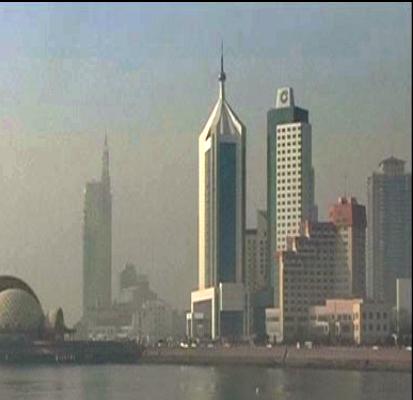}
		\captionsetup{labelformat=empty}
		\captionsetup{justification=centering}
	\end{minipage} 
	\begin{minipage}{.135\textwidth}
		\centering
		\includegraphics[width=2.4cm, height=2cm]{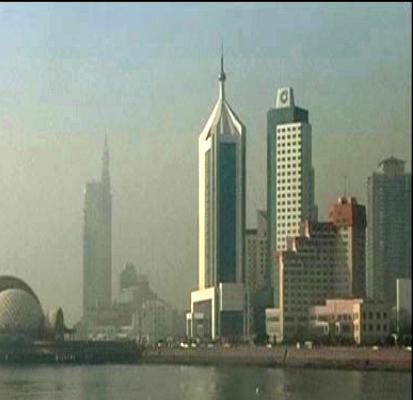}
		\captionsetup{labelformat=empty}
		\captionsetup{justification=centering}
	\end{minipage} 
	\begin{minipage}{.135\textwidth}
		\centering
		\includegraphics[width=2.4cm, height=2cm]{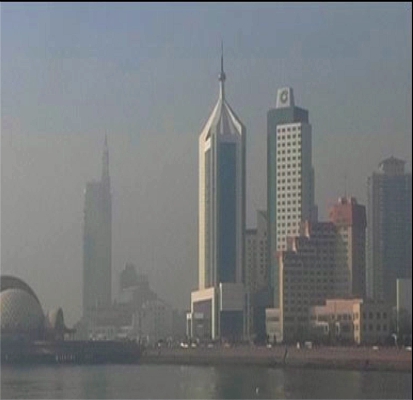}
		\captionsetup{labelformat=empty}
		\captionsetup{justification=centering}
	\end{minipage}
	\begin{minipage}{.135\textwidth}
		\centering
		\includegraphics[width=2.4cm, height=2cm]{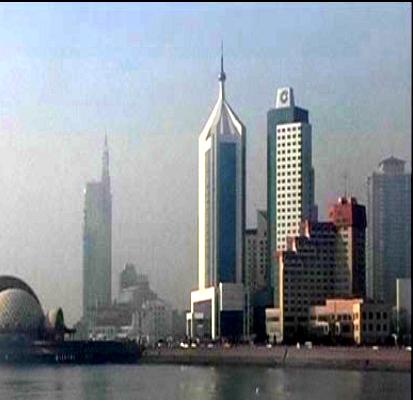}
		\captionsetup{labelformat=empty}
		\captionsetup{justification=centering}
	\end{minipage} \\		\vskip+4pt
	\begin{minipage}{.135\textwidth}
		\centering
		\includegraphics[width=1\textwidth]{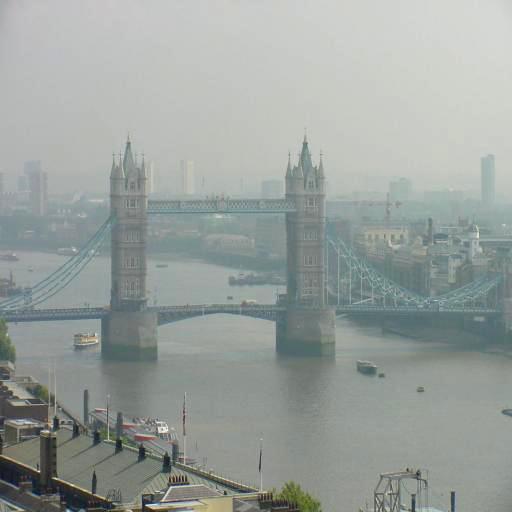}
		\captionsetup{labelformat=empty}
		\captionsetup{justification=centering}
	\end{minipage} 
	\begin{minipage}{.135\textwidth}
		\centering
		\includegraphics[width=1\textwidth]{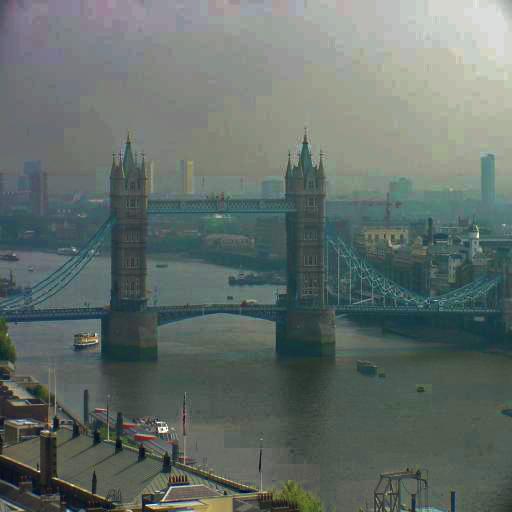}
		\captionsetup{labelformat=empty}
		\captionsetup{justification=centering}
	\end{minipage} 
	\begin{minipage}{.135\textwidth}
		\centering
		\includegraphics[width=1\textwidth]{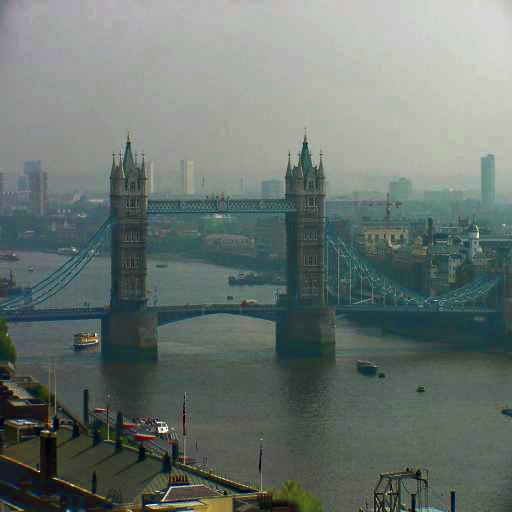}
		\captionsetup{labelformat=empty}
		\captionsetup{justification=centering}
	\end{minipage} 
	\begin{minipage}{.135\textwidth}
		\centering
		\includegraphics[width=1\textwidth]{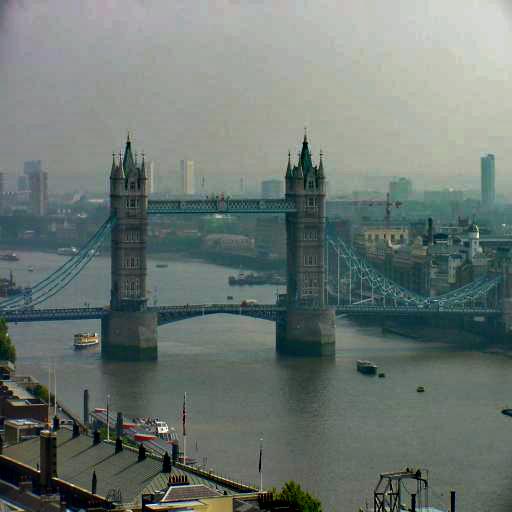}
		\captionsetup{labelformat=empty}
		\captionsetup{justification=centering}
	\end{minipage} 
	\begin{minipage}{.135\textwidth}
		\centering
		\includegraphics[width=1\textwidth]{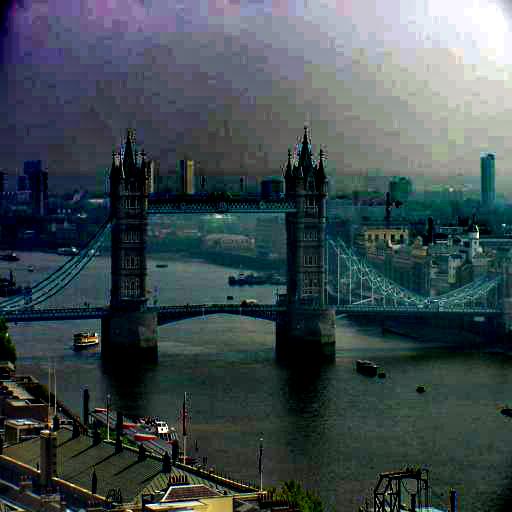}
		\captionsetup{labelformat=empty}
		\captionsetup{justification=centering}
	\end{minipage} 
	\begin{minipage}{.135\textwidth}
		\centering
		\includegraphics[width=1\textwidth]{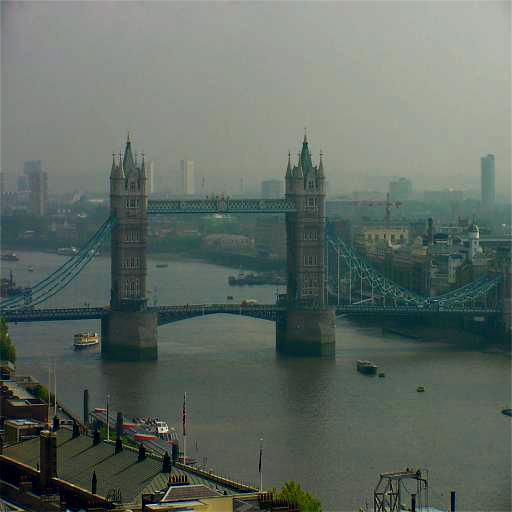}
		\captionsetup{labelformat=empty}
		\captionsetup{justification=centering}
	\end{minipage}
	\begin{minipage}{.135\textwidth}
		\centering
		\includegraphics[width=1\textwidth]{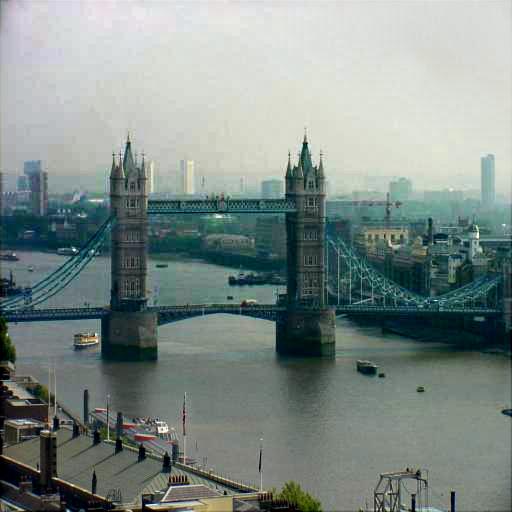}
		\captionsetup{labelformat=empty}
		\captionsetup{justification=centering}
	\end{minipage} \\		\vskip+4pt
	\begin{minipage}{.135\textwidth}
		\centering
		\includegraphics[width=2.4cm, height=1.5cm]{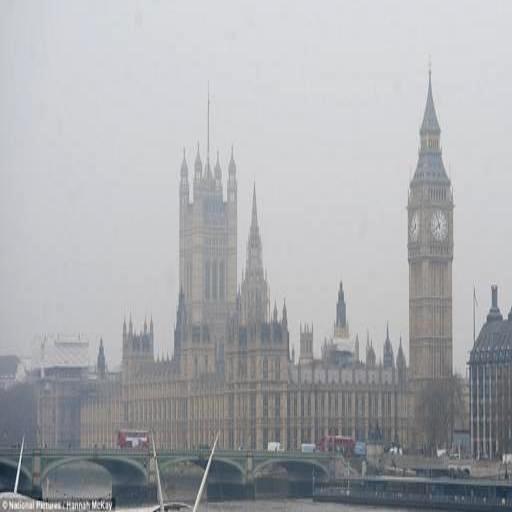}
		\captionsetup{labelformat=empty}
		\captionsetup{justification=centering}
	\end{minipage} 
	\begin{minipage}{.135\textwidth}
		\centering
		\includegraphics[width=2.4cm, height=1.5cm]{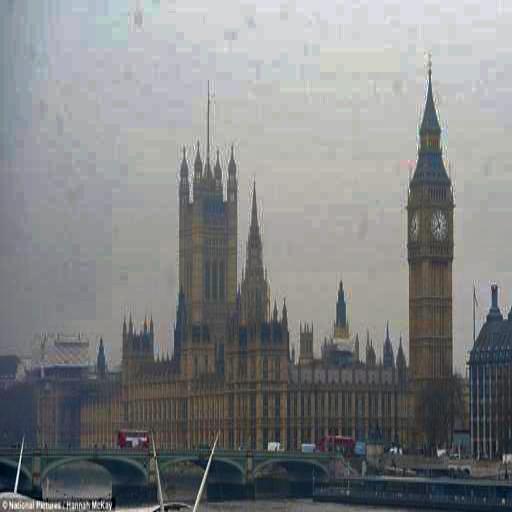}
		\captionsetup{labelformat=empty}
		\captionsetup{justification=centering}
	\end{minipage} 
	\begin{minipage}{.135\textwidth}
		\centering
		\includegraphics[width=2.4cm, height=1.5cm]{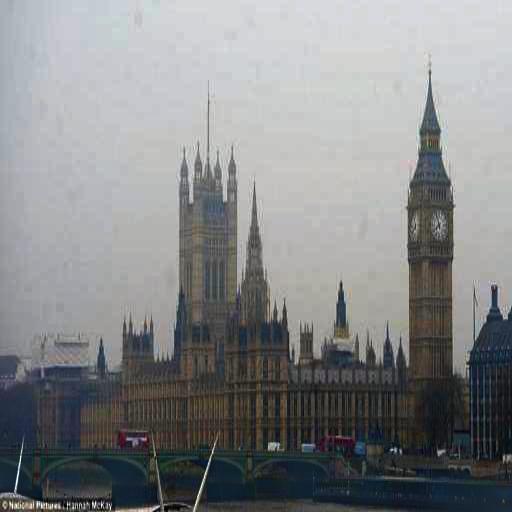}
		\captionsetup{labelformat=empty}
		\captionsetup{justification=centering}
	\end{minipage} 
	\begin{minipage}{.135\textwidth}
		\centering
		\includegraphics[width=2.4cm, height=1.5cm]{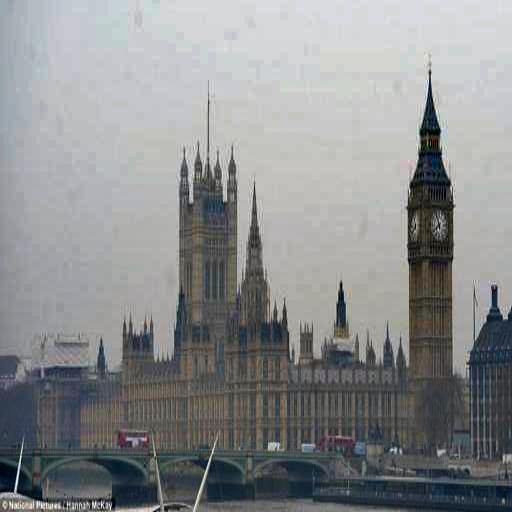}
		\captionsetup{labelformat=empty}
		\captionsetup{justification=centering}
	\end{minipage}
	\begin{minipage}{.135\textwidth}
		\centering
		\includegraphics[width=2.4cm, height=1.5cm]{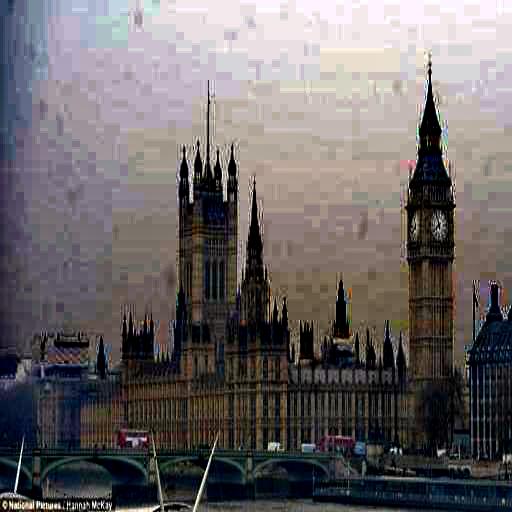}
		\captionsetup{labelformat=empty}
		\captionsetup{justification=centering}
	\end{minipage} 
	\begin{minipage}{.135\textwidth}
		\centering
		\includegraphics[width=2.4cm, height=1.5cm]{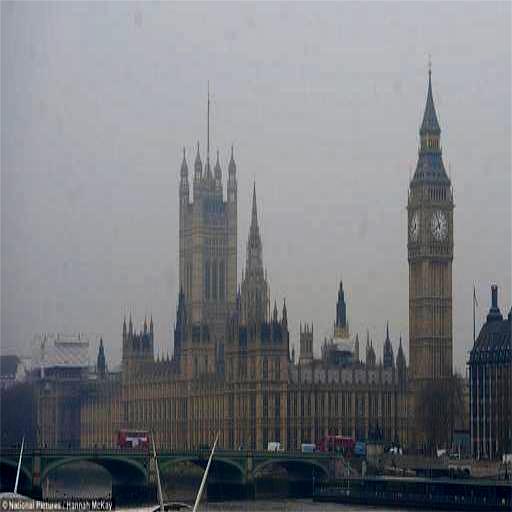}
		\captionsetup{labelformat=empty}
		\captionsetup{justification=centering}
	\end{minipage}
	\begin{minipage}{.135\textwidth}
		\centering
		\includegraphics[width=2.4cm, height=1.5cm]{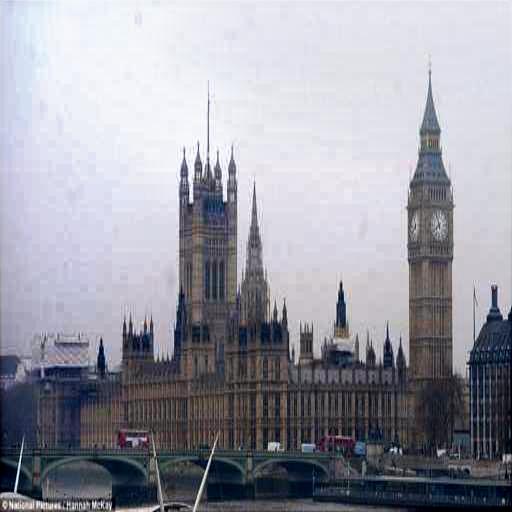}
		\captionsetup{labelformat=empty}
		\captionsetup{justification=centering}
	\end{minipage} \\		\vskip+7pt
	\begin{minipage}{.135\textwidth}
		\centering
		\includegraphics[width=2.4cm, height=3.99cm]{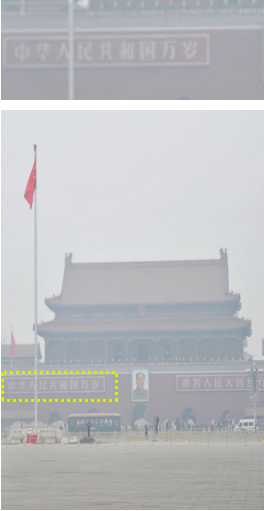}
		\captionsetup{labelformat=empty}
		\captionsetup{justification=centering}
		\caption*{Input\\\quad}
	\end{minipage} 
	\begin{minipage}{.135\textwidth}
		\centering
		\includegraphics[width=2.4cm, height=3.99cm]{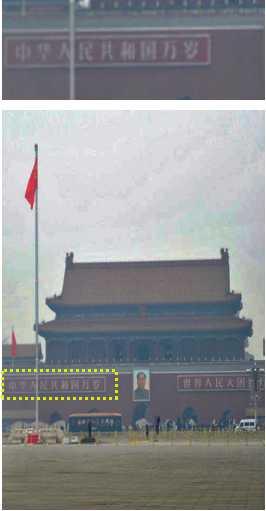}
		\captionsetup{labelformat=empty}
		\captionsetup{justification=centering}
		\caption*{He. \emph{et al.} (CVPR'09) \cite{dehaz_2009_dark}}
	\end{minipage} 
	\begin{minipage}{.135\textwidth}
		\centering
		\includegraphics[width=2.4cm, height=3.99cm]{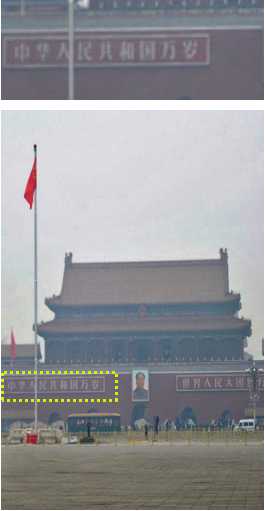}
		\captionsetup{labelformat=empty}
		\captionsetup{justification=centering}
		\caption*{Zhu. \emph{et al.} (TIP'15) \cite{dehaze_fast15}}
	\end{minipage} 
	\begin{minipage}{.135\textwidth}
		\centering
		\includegraphics[width=2.4cm, height=3.99cm]{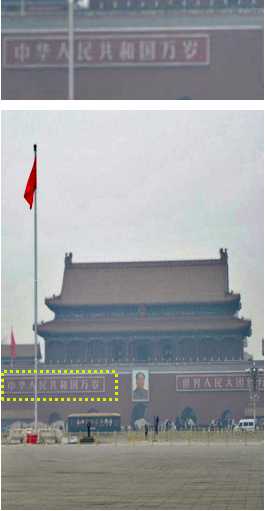}
		\captionsetup{labelformat=empty}
		\captionsetup{justification=centering}
		\caption*{Ren. \emph{et al.} (ECCV'16) \cite{dehaze_2016_eccv}}
	\end{minipage} 
	\begin{minipage}{.135\textwidth}
		\centering
		\includegraphics[width=2.4cm, height=3.99cm]{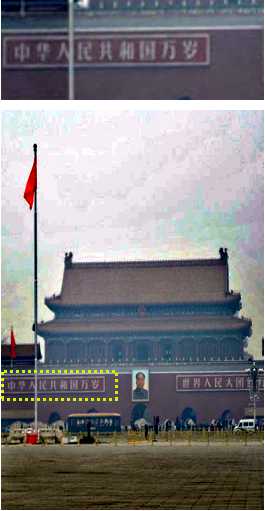}
		\captionsetup{labelformat=empty}
		\captionsetup{justification=centering}
		\caption*{Berman \emph{et al.} (CVPR'16) \cite{dehaze_2016_dana,dehaze_iccp217}}
	\end{minipage} 
	\begin{minipage}{.135\textwidth}
		\centering
		\includegraphics[width=2.4cm, height=3.99cm]{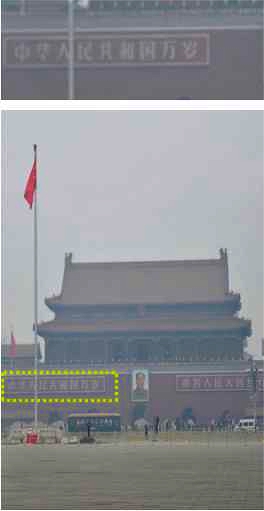}
		\captionsetup{labelformat=empty}
		\captionsetup{justification=centering}
		\caption*{Li. \emph{et al.} (ICCV'17) \cite{dehaze_iccv17} }
	\end{minipage}
	\begin{minipage}{.135\textwidth}
		\centering
		\includegraphics[width=2.4cm, height=3.99cm]{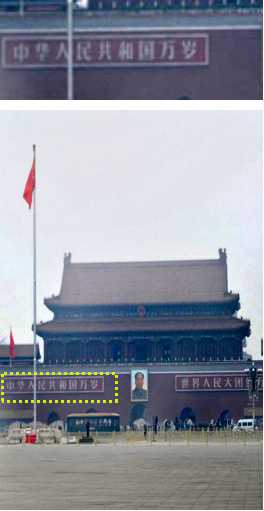}
		\captionsetup{labelformat=empty}
		\captionsetup{justification=centering}
		\caption*{DCPDN\\\qquad}
	\end{minipage} 
	\vskip -4pt\caption{Dehazing results evaluated on real-world images downloaded from the Internet.}\label{nat_haze2}
\vskip -8pt
\end{figure*}

\subsection{Comparison with state-of-the-art Methods}
To demonstrate the improvements achieved by the proposed method, it is compared against the recent state-of-the-art methods \cite{dehaz_2009_dark,dehaze_fast15,dehaze_2016_eccv,dehaze_2016_dana,dehaze_iccp217,dehaze_iccv17}. on both synthetic and real datasets. 

\noindent \textbf{Evaluation on synthetic dataset:} The proposed network is  evaluated on two synthetic datasets \textbf{TestA} and \textbf{TestB}. Since the datasets are synthesized, the ground truth images and the transmission maps are available, enabling us to evaluate the performance qualitatively as well as quantitatively.   Sample results for the proposed method and five recent state-of-the-art methods, on two sample images from the test datasets are shown in Fig.~\ref{fig:synthetic}. It can be observed that even though previous methods are able to remove haze from the input image, they tend to either over dehaze or under dehaze the image making the result darker or leaving some haze in the result. In contrast, it can be observed from our results that they preserve sharper contours with less color distortion and are more visually closer to the ground-truth. The quantitative results, tabulated in Table \ref{ta:quantitive_depth1} and Table~\ref{ta:quantitive_depth2} \footnote{N/A: Code released is unable to estimate the transmission map.  }, evaluated on both \textbf{TestA} and \textbf{TestB} also demonstrate the effectiveness of the proposed method.

\noindent \textbf{Evaluation on a real dataset:} To demonstrate the generalization ability of the proposed method, we evaluate the proposed method on several real-world hazy images provided by previous methods and other challenging hazy images  downloaded from  the Internet. 

Results for four sample images obtained from the previous methods \cite{dehaze_2016_eccv,dehaze_2016_tip,dehaze_2014acm} are shown in Fig.~\ref{fig:nature_old}. As revealed in  Fig.~\ref{fig:nature_old},  methods of He \emph{et al.} \cite{dehaz_2009_dark} and Ren \emph{et al.} \cite{dehaze_2016_eccv} (observed on the fourth row)  tend to leave haze in the results and methods of Zhu \emph{et al.} \cite{dehaze_fast15} and Li \emph{et al.} \cite{dehaze_iccv17}(shown on the second row)  tend to darken some regions (notice the background wall).
Methods from Berman \emph{et al.} \cite{dehaze_2016_dana,dehaze_iccp217} and our method have the most competitive visual results. However, by looking closer, we observe that Berman \emph{et al.} \cite{dehaze_2016_dana,dehaze_iccp217} produce unrealistic color shifts such as the building color in the fourth row. In contrast, our method is able to generate realistic colors while better removing haze.  This can be seen by comparing the first and the second row.

We also evaluate on several hazy images downloaded from the Internet. The dehazed results are shown in Fig.~\ref{nat_haze2}. It can be seen from these results that outputs from He \emph{et al.} \cite{dehaz_2009_dark} and Berman \emph{et al.} \cite{dehaze_2016_dana,dehaze_iccp217} suffer from color distortions, as shown in the second and third rows.  In contrast, our method is able to achieve better dehazing with visually appealing results.

\section{Conclusion}
We  presented a new end-to-end deep learning-based  dehazing method that can jointly optimize transmission map, atmospheric light and dehazed image. This is achieved via directly embedding the atmospheric image degradation model into the overall optimization framework. To efficiently estimate the transmission map, a novel densely connected encoder-decoder structure with multi-level pooling module is proposed and this network is optimized by a new edge-preserving loss. In addition, to refine the details and to leverage the mutual structural correlation between the dehazed image and the estimated transmission map, a joint-discriminator based GAN framework is introduced in the proposed method.  Various experiments were conducted to show the significance of the proposed method.

\section*{Acknowledgement}
 This work was supported by an ARO grant W911NF-16-1-0126.

{\small
\bibliographystyle{ieee}
\bibliography{egbib}
}

\end{document}